\documentclass{article}

\usepackage{arxiv}
\usepackage{bm}
\usepackage[utf8]{inputenc} % allow utf-8 input
\usepackage[T1]{fontenc}    % use 8-bit T1 fonts
\usepackage{hyperref}       % hyperlinks
\usepackage{url}            % simple URL typesetting
\usepackage{booktabs}       % professional-quality tables
\usepackage{amsfonts}       % blackboard math symbols
\usepackage{nicefrac}       % compact symbols for 1/2, etc.
\usepackage{microtype}      % microtypography
\usepackage{lipsum}
\usepackage{graphicx}
\graphicspath{ {./images/} }
\usepackage[sort,numbers]{natbib}
\usepackage{subcaption}
\usepackage{comment}
\usepackage{mathtools}
\usepackage{xcolor}
\usepackage{amsmath,amssymb}
\usepackage{gensymb}
\usepackage{mathrsfs}
\usepackage{multirow}
\usepackage{tikz}
\newcommand*\circled[1]{\tikz[baseline=(char.base)]{
            \node[shape=circle,draw,inner sep=1pt] (char) {#1};}}
            
\title{SkyGPT: Probabilistic Short-term Solar Forecasting Using Synthetic Sky Videos from Physics-constrained VideoGPT}

\author{
 Yuhao Nie\thanks{Corresponding author}\\
  Energy Science and Engineering\\
  Stanford University\\
  United States \\
  \texttt{ynie@stanford.edu} \\
   \And
 Eric Zelikman\footnotemark[2]\\
  Computer Science\\
  Stanford University\\
  United States \\
  \texttt{ezelikman@cs.stanford.edu} \\
  \And
 Andea Scott\thanks{The two authors have equal contributions.}\\
  Energy Science and Engineering\\
  Stanford University\\
  United States \\
  \texttt{andea98@stanford.edu} \\
  \And
 Quentin Paletta \\
  Department of Engineering\\
  University of Cambridge\\
  United Kingdom \\
  \texttt{qp208@cam.ac.uk} \\
  \AND
  Adam Brandt \\
  Energy Science and Engineering\\
  Stanford University\\
  United States \\
  \texttt{abrandt@stanford.edu} \\
}

\begin{document}
\maketitle
\begin{abstract}
In recent years, deep learning-based solar forecasting using all-sky images has emerged as a promising approach for alleviating uncertainty in PV power generation. However, the stochastic nature of cloud movement remains a major challenge for accurate and reliable solar forecasting. With the recent advances in generative artificial intelligence, the synthesis of visually plausible yet diversified sky videos has potential for aiding in forecasts. In this study, we introduce \emph{SkyGPT}, a physics-informed stochastic video prediction model that is able to generate multiple possible future images of the sky with diverse cloud motion patterns, by using past sky image sequences as input. Extensive experiments and comparison with benchmark video prediction models demonstrate the effectiveness of the proposed model in capturing cloud dynamics and generating future sky images with high realism and diversity. Furthermore, we feed the generated future sky images from the video prediction models for 15-minute-ahead probabilistic solar forecasting for a 30-kW roof-top PV system, and compare it with an end-to-end deep learning baseline model SUNSET and a smart persistence model. Better PV output prediction reliability and sharpness is observed by using the predicted sky images generated with SkyGPT compared with other benchmark models, achieving a continuous ranked probability score (CRPS) of 2.81 (13\% better than SUNSET and 23\% better than smart persistence) and a Winkler score of 26.70 for the test set. Although an arbitrary number of futures can be generated from a historical sky image sequence, the results suggest that 10 future scenarios is a good choice that balances probabilistic solar forecasting performance and computational cost. 
\end{abstract}

% keywords can be removed

\section{Introduction}
\label{sec:intro}
Renewable energy sources, such as solar photovoltaics (PV), will be the key component for future power systems \cite{IPCC2011}. One of the challenges for large-scale PV integration is unstable power generation. PV output can greatly fluctuate on short time horizons due to cloud passage events \cite{Sun2019_thesis}. Although this variability can be filled in by dispatchable resources, such as gas turbines, in the current grid, it will become a critical issue as transition to a high-renewable energy system continues. Accurate short-term solar forecasting systems are therefore needed to reduce the uncertainty in PV power generation, help grid operators optimize assets and dispatch priorities, and minimize mitigation costs (e.g., batteries).

\setcounter{footnote}{0} 
Ground-based sky images have emerged as a promising approach to observe the surrounding cloud cover for short-term\footnote{Although as yet there is no common agreement on the classification criterion, we use the definition of forecasting horizon less than 30 minutes in this study for short-term solar forecasting \cite{van2018review}.} solar forecasting \cite{chow2011intra,marquez2013intra} due to their high temporal (from seconds to minutes) and spatial resolution ($<$1x kms) \cite{van2018review}. Satellite imagery and numerical weather prediction, on the other hand, with coarse temporal (from minutes to 10x hours for satellite, from minutes up to 1000x hours for NWP) and spatial resolution (1x$\sim$100x kms) \cite{van2018review}, fit better for medium- to long-term forecasting at a scale of a few hours to a day ahead.

Traditional image-based forecasting methods focused on handcrafted feature engineering of sky images. Extracted features, such as red-blue ratio, cloud coverage, and cloud motion vectors, are used for building physical deterministic models \cite{chow2011intra,marquez2013intra,quesada-ruizCloudtrackingMethodologyIntrahour2014a} or training machine learning models \cite{chuHybridIntrahourDNI2013a,Chu2015realtime,Chu2015reforcast,pedroAdaptiveImageFeatures2019}. In the past five years, with the development of computer vision techniques, efforts have shifted to build end-to-end deep learning models that correlate the future PV ouput (or solar irradiance) with the corresponding historical sky image sequence as well as other auxiliary input such as the past PV output (or solar irradiance), sun angles and/or weather data. These deep learning models often rely on convolutional neural networks (CNNs), either using CNNs solely \cite{Sun2019,Nie2020,Feng2020,palettaConvolutionalNeuralNetworks2020,nie2021resampling,feng2022convolutional} or combining CNNs with recurrent neural networks (RNNs), like long short-term memory (LSTM) \cite{Zhang2018,palettaBenchmarkingDeepLearning2021,Paletta2021eclipse}.

Existing deep learning-based solar forecasting models often suffer from temporal lags in prediction \cite{Sun2019,palettaBenchmarkingDeepLearning2021}, especially on cloudy days when the power output of a PV system can drastically change within a short time (so-called ramp events). This indicates that the models tend to rely on past observations to make predictions and that cloud dynamics are not well captured. Predicting cloud motion is very challenging as clouds can continuously deform and condense or evaporate during movement \cite{LeGuen2020_solar}. These stochastic behaviors of clouds are the main cause of uncertainty of PV panel output. However, the complex physics involved in cloud propagation is hard to learn via the end-to-end training of solar forecasting models. One approach is to use a dedicated cloud motion prediction model to better capture the cloud dynamics. Good cloud motion predictions should not only capture the general trend of cloud movement, but also account for the stochasticity. Traditional image-based motion prediction methods, such as particle image velocimetry \cite{willert1991digital} and optical flow \cite{beauchemin1995computation}, which conduct patch matching by assuming linear propagation of objects, are challenged by the complexity of clouds. 

Another issue with the current solar forecasting research is that prediction uncertainty is seldom quantified. Most existing studies focus on deterministic prediction \cite{van2018review}, i.e., predicting a single value of either future PV power output or solar irradiance. Limited studies have investigated probabilistic solar power forecasting, i.e., generating a range prediction covering the uncertainty of future power generation, which is more valuable for grid risk management. Existing probabilistic solar forecasting efforts include, for example, using bootstrap sampling for training multiple artificial neural network models \cite{Chu2015realtime}, training neural networks to predict lower and upper bound of PV power to generate prediction intervals \cite{ni2017ensemble}, natural gradient boosting methods with posthoc calibration \cite{zelikman2020short}, and discretizing the target space into bins (e.g., discretizing irradiance with N equally spaced bins from 0 to 1300 W/m$^2$) and predicting the probability of future values falling in each bin \cite{Paletta2021eclipse,Paletta2023omnivision}. However, very little work has explored estimating the uncertainty in power generation based on the realizations of different possible future sky conditions due to the challenges in capturing the stochastic cloud dynamics. \looseness=-1

Recent improvements in generative artificial intelligence, specifically the advances in image and video synthesis, has provided opportunities for tackling these challenges. Future sky image frames can be generated given a set of past sky image frames as input, based on the underlying cloud motion patterns learned by the video prediction model during training. By leveraging the recent advances in video prediction, we propose in this study a two-stage deep learning-based probabilistic solar forecasting system. This forecasting system uses a physics-constrained stochastic sky video predictor that is capable of generating a range of possible future sky videos from the same historical sky image sequence, followed by a CNN-based PV output predictor trained to generate a range of predictions of the future PV output based on the predicted future images. Our contributions are summarized as follows: 
\begin{enumerate}
    \item We developed a specialized stochastic video prediction framework that can capture the physical dynamics in the context of sky videos and generate visually plausible yet remarkably diversified videos of the future sky.
    \item We qualitatively and quantitatively examined and compared different benchmark deep video prediction models for generating future sky videos.
    \item We applied the predicted frames for probabilistic solar forecasting tasks and showed the promise of using these synthesized images for uncertainty estimation in PV output prediction.
\end{enumerate}

The rest of this paper is organized as follows: Section \ref{sec:cloud_motion_prediction_review} provides an overview of deep video prediction models. 
Section \ref{sec:methodology} presents the proposed methodology, including the model architectures, training details,  a baseline model for comparison, and evaluation metrics. Section \ref{sec:dataset} describes the dataset used for the experiments, and Section \ref{sec:experiments} analyzes the results for both video prediction and probabilistic solar forecasting. Section \ref{sec:limitations} discusses some limitations of this study and provides directions for future research. Finally, we summarize the findings of this study in Section \ref{sec:conclusion}.

\section{Overview of Video Prediction Models}
\label{sec:cloud_motion_prediction_review}

A video stream is made up of individual frames, each one representing a time slice of the scene. The goal of video prediction is to generate plausible future frames given a set of historical frames. The dynamics from the history are captured and extrapolated into the future based on the underlying patterns learned from the training data. Deep networks such as CNNs, RNNs and generative models commonly serve as the backbone for video prediction models \cite{Oprea2020}. In recent years, there has been a trend of replacing RNNs with transformers \cite{vaswani2017attention} in model architectures for better handling of long-term dependencies in sequential data.

Although the existing video prediction models vary in architecture, they can be divided into two categories based on the prediction type, i.e., deterministic models and stochastic models. Deterministic video prediction models generate only a single future given one set of historical inputs, while their stochastic counterparts can generate multiple possible futures from the same historical input. Deterministic video prediction models can extrapolate the frames in the immediate future with high precision, but they struggle when making long-term predictions in multi-modal natural videos, e.g., sky videos with stochastic cloud motion. To accommodate uncertainty, they tend to average out plausible future outcomes into a single blurry prediction \cite{Oprea2020}. This behavior is due to the pixel-wise losses used in model training, e.g., cross-entropy and mean-squared error. Nevertheless, the techniques developed in deterministic video prediction for capturing the scene dynamics can be applied in stochastic video prediction methods.

Early video prediction work focused on extending classical RNNs to more sophisticated recurrent models to deal with the long-term spatiotemporal dependencies of video sequences. Most of these methods are deterministic. For example, \citet{Shi2015} incorporated convolution operation into the original LSTM module (so-called ConvLSTM) and extended the use of LSTM-based models to the image space. ConvLSTM was originally proposed for precipitation nowcasting using radar echo maps, but it has later become a building block for many benchmark video prediction models. For example, \citet{Wang2017_predrnn} improved the memory flows in ConvLSTM module to deal with both short and long-term dynamics in the spatiotemporal sequence data. \citet{Wang2019_mim} further decomposed the stationary and non-stationary properties in spatiotemporal dynamics to handle the higher-order non-stationarity. \citet{LeGuen2020} proposed a two-branch deep architecture PhyDNet that disentangles the physics dynamics and the residual information (e.g., appearance, texture, details) of the video in the latent space. The physics dynamics were modeled by a physically-constrained recurrent cell called PhyCell, and the residual information was learned by ConvLSTM.

Recently, deep generative models (DGMs) have seen increasing popularity in video prediction, featured by two most commonly used and efficient approaches, namely, variational auto-encoders (VAEs) \cite{kingma2013auto} and generative adversarial networks (GANs) \cite{Ian2014}. DGMs are statistical models that aim at learning probability distributions approximating distributions of the input data. As DGMs are probabilistic models, new samples can be easily generated from the learned distributions. However, it should be noted that VAEs and GANs have their own strengths and weaknesses. VAEs, which learn probability distributions of the input data explicitly, can be used to generate diverse futures by sampling latent variables, while their predictions can be blurry as they still apply a pixel-wise MSE loss. GANs, on the other hand, can generate very realistic and sharp images via adversarial training, but without an explicit latent variable, they typically work with deterministic models or incorporate stochasticity only through input noise which has limited capability of generating diverse futures \cite{Lee2018}.

\citet{ravuri2021skilful} used GANs for precipitation nowcasting by simulating many samples from the conditional distribution of future radar given historical radar and generating a collection of forecasts. The stochasticity of their generations comes from injecting Gaussian noise into the input radar data. \citet{Lee2018} combined the strengths of GANs and VAEs and built a Stochastic Adversarial Video Prediction (SAVP) model that managed to generate future frames with both fidelity and diversity. Similarly, a model named VideoGPT was proposed by \citet{Yan2021} to address the stochasticity and realistic future image generation issues in video prediction. The model is featured by a Vector Quantized-VAE (VQ-VAE) that learns a discretized latent from the input videos and a transformer that autoregressively model the discrete latents. VideoGPT achieved promising performance on various video prediction benchmark datasets, including Moving MNIST \cite{srivastava2015unsupervised}, BAIR Robot Pushing \cite{ebert2017} and UCF-101 \cite{soomro2012ucf101}, and it demonstrated the capability of generating highly diverse futures even just given one historical frame as input.

Limited studies have applied video prediction models to cloud motion prediction. Most of the existing studies are deterministic without accounting for the stochasticity of cloud motion, and very few of them have examined using the prediction results for solar forecasting. \citet{andrianakos2019sky} developed a deep convolution GAN (DCGAN) architecture for future sky images forecasting and further using the predicted images for cloud coverage estimation. It showed the effectiveness of using GANs to generate realistic and sharp future sky images. However, minimal gains were observed using the sharp images generated by GANs for cloud coverage estimation compared with the blurry images generated by the same generator architecture without adversarial training. \citet{LeGuen2020_solar} used the modified PhyDNet architecture for extracting spatiotemporal features from historical sky image sequences to forecast sky images and solar irradiance at the same time for up to 5 minutes ahead. It was claimed to outperform strong baselines in multiple performance metrics, including mean squared error mean absolute error and structural similarity index. Ground-based sky images taken by fish-eye cameras or total sky imagers have distortions, particularly at the horizon, due to their wide field-of-view. \citet{JulianSpatialWarping} examined the effectiveness of spatial warping for forecasting sky images and found the wrapped sky images can facilitate longer forecasting of cloud evolution.

\section{Methodology}
\label{sec:methodology}

%\subsection{Proposed Solar Forecasting Framework}
In this study, we focus on short-term probabilistic solar forecasting, with a specific aim of generating range predictions of PV power output 15 minutes ahead into the future. We propose a two-stage deep learning-based probabilistic forecasting framework to address the existing challenges in solar forecasting, i.e., poor modeling of cloud dynamics and the lack of uncertainty quantification in PV output prediction. 

\begin{figure}[h!]
    \centering
    \includegraphics[width=1.\textwidth]{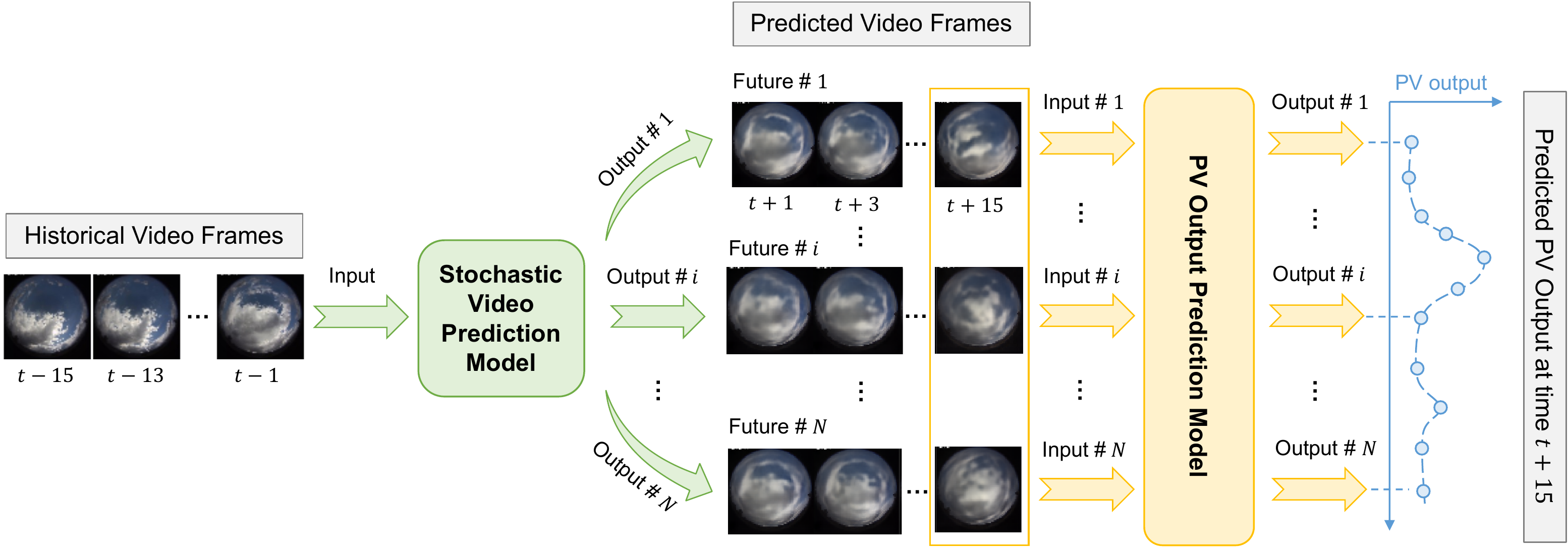}
    \caption{Proposed probabilistic solar forecasting framework. Historical video frames from the past 15 minutes with time stamps 2 minutes apart from each other are used as input to predict the next 15 minutes' future video frames with the same temporal resolution. Only the last predicted frames (at time $t+15$) are used for PV output prediction.}
    \label{fig:solar_forecasting_system}
\end{figure}

The proposed framework is made up of a stochastic sky video predictor, which is capable of generating visually plausible and diversified future sky videos that captures the dynamics from the historical sky image sequence, followed by a PV output predictor, which maps the synthetic future sky images to concurrent PV output for generating range predictions of the future PV output. A schematic of the proposed forecasting system is shown in Figure \ref{fig:solar_forecasting_system}. It should be noted that only the last predicted sky image frames are utilized for PV output prediction and the proposed system does not take in the historical PV output as input, preventing the model from overfitting to this signal. %More details can be found in the following sections.

%the future images frames are generated in an iterative fashion (i.e., a new frame is generated based on the previously predicted frames) and 

\subsection{Stochastic Sky Video Prediction}

\label{subsec:framework_overview}
The sky video prediction problem is formulated as a sequence-to-sequence forecasting problem. A video prediction model is developed to predict future sky images up to 15 minutes ahead, from time $t+1$ to $t+15$ with a 2-minute sampling interval (i.e., samples generated for $t+1$, $t+3$, $t+5$,...). These are generated given a historical image sequence from the past 15 minutes, from $t-15$ to $t-1$ with the same sampling interval. For stochastic prediction, the model is able to generate multiple possible futures conditioned on the same historical inputs via sampling different latent sequences from the prior. More details can be found in the Model Architecture below.

\paragraph{Model Architecture}
\label{para:model_arch}

We name our proposed stochastic sky video prediction model \emph{SkyGPT}, which is inspired by two emerging video prediction models VideoGPT \cite{Yan2021} and PhyDNet \cite{LeGuen2020}. VideoGPT is a stochastic video prediction model that is capable of generating realistic samples competitive with state-of-the-art GAN models and also shows remarkable performance in generating divergent images from the same inputs. In addition, the use of a transformer architecture enables it to effectively model long-term spatiotemporal dependencies. However, VideoGPT has not been examined for cloud motion prediction, with challenges arising from the volatility and stochasticity of clouds. PhyDNet is a deterministic RNN-based architecture that incorporates physical knowledge, represented by linear partial differential equations (PDEs), for modeling the physics dynamics in video. PhyDNet has been applied for cloud motion prediction by \citet{LeGuen2020_solar} and demonstrates effectiveness in capturing the general trend of cloud motion, but the generated images are quite blurry even for 5-minute-ahead forecast and without any stochasticity given the deterministic property of the model. The combination of the above two architectures could be complementary and potentially provide benefits for the 15-minute-ahead cloud motion forecast problem that we are trying to tackle in this study. 

The SkyGPT follows the general structure of VideoGPT, which consists of two main parts, a vector quantized variational auto-encoder (VQ-VAE) \cite{Oord2017} and an image transformer \cite{chen2020generative}. The VQ-VAE encompasses an encoder-decoder architecture similar to classical VAEs, but it learns a discrete latent representation of input data instead of a continuous one.  The encoder part ($E$) consists of a series of 3D convolutions that downsample over space-time followed by attention residual blocks \cite{Yan2021}, to compress high dimensional input video data ($x$) into latent vectors ($h$). The latent vectors are then discretized by performing a nearest neighbors lookup in a codebook of embeddings $C = \{e_i| 1\leq i\leq K\}$, where $K$ is an adjustable parameter representing the size of the codebook. These embeddings are initialized randomly and can be learned during the training of the model. The decoder part ($D$) has a reverse architectural design as the encoder, with attention residual blocks followed by a series of 3D transposed convolutions that upsample over space-time to reconstruct the input videos from the quantized embeddings. The image transformer, as a prior network, is used to model the latent tokens in an auto-regressive fashion, where new predictions are made by feeding back the predictions from previous steps. The auto-regressive modeling is performed in the downsampled latent space rather than the raw image pixel space to avoid spatiotemporal redundancies in high dimensional imagery data \cite{Yan2021}. The generated latents from the transformer are then decoded to videos of the original resolution using the decoder of the VQ-VAE. In our case, future sky image generation is conditioned on historical sky image frames. Thus, a conditional prior network is trained by first feeding the conditional frames from time $t-15$ to $t-1$ into a 3D ResNet and then performing cross-attention on the ResNet output representation during network training as \citet{Yan2021} did for VideoGPT. With the learned conditional prior, diversified futures can be sampled from the latent conditioned on the historical frames during inference.

\begin{figure}[h!]
\centering
\includegraphics[width=.85\textwidth]{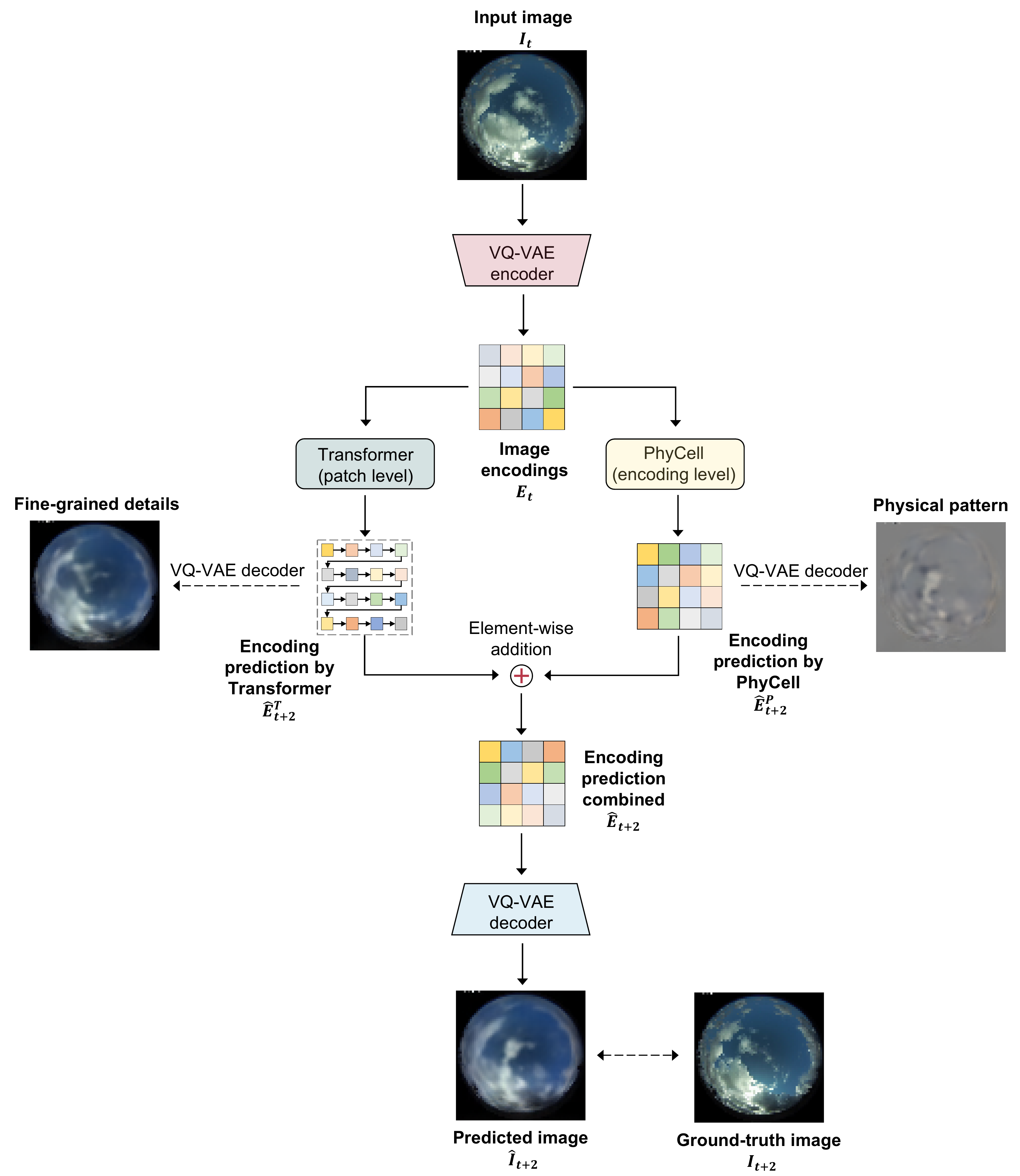}
\caption{SkyGPT for future sky image prediction. The image at time $t$ is used to predict the image at the next time step $t+2$. The prediction is disentangled in the encoding space by the PhyCell and Transformer. For visualization purposes, the next step encodings predicted by PhyCell and Transformer are decoded, which shows that PhyCell captures the physical pattern of the motion, while Transformer is responsible for filling in the prediction with fine-grained details.}
\label{fig:SkyGPT_architecture}
\end{figure}

To enhance the modeling of cloud motion, we incorporate prior physical knowledge into the transformer by adapting a PDE-constrained module called PhyCell from the PhyDNet \cite{LeGuen2020} for latent modeling. We call this entire architecture a Phy-transformer (in short of physics-informed transformer) to distinguish it from the transformer component within the architecture. Specifically, the latent modeling is disentangled into two branches here, i.e., the modeling of physics dynamics fulfilled by the PhyCell and the modeling of fine-grained details fulfilled by the transformer. The transformer works at the patch level while the PhyCell works at the image embedding level, and the predicted embeddings from the two branches are finally combined and decoded into the predicted frames of the original resolution. PhyCell implements a two-step scheme, a physical prediction with convolutions for approximating the spatial derivatives of a generic class of linear PDEs, which represent a wide range of classical physical models, e.g., the heat equation, the wave equations, the advection-diffusion equations, followed by an input assimilation as a correction of latent physical dynamics driven by observed data. An illustration of the model can be found in Figure \ref{fig:SkyGPT_architecture}. %More architecture details can be found in Appendix \ref{appendixA:model_arch_details}.

\paragraph{Objective Functions} 

The VQ-VAE and Phy-transformer are trained separately. First, the VQ-VAE is trained with the following objective function \cite{Yan2021}:

\begin{equation}\label{vqvae_loss_full}
    \mathcal{L}_{VQ-VAE} = \mathcal{L}_{recon}+ \mathcal{L}_{codebook}+ \beta\mathcal{L}_{commit}\\
\end{equation}
with 
\begin{equation}\label{vqvae_loss_recon}
    \mathcal{L}_{recon} = \|x-D(e)\|_2^2
\end{equation}
\begin{equation}\label{vqvae_loss_codebook}
    \mathcal{L}_{codebook} = \|sg[E(x)]-e\|_2^2
\end{equation}
\begin{equation}\label{vqvae_loss_commit}
    \mathcal{L}_{commit} = \|sg[e]-E(x)\|_2^2
\end{equation}

Where $sg$ refers to a stop-gradient. The objective consists of a reconstruction loss $\mathcal{L}_{recon}$, a codebook loss $\mathcal{L}_{codebook}$, and a commitment loss $\mathcal{L}_{commit}$. The reconstruction loss encourages the VQ-VAE to learn good representations to accurately reconstruct data samples. The codebook loss brings codebook embeddings closer to their corresponding encoder outputs, and the commitment loss is weighted by a hyperparameter $\beta$ and prevents the encoder outputs from fluctuating between different code vectors.

The objective function for training the Phy-transformer contains two parts, the moment loss $\mathcal{L}_{moment}$ from the PhyDNet PhyCell, which enforces the convolution operations to approximate the spatial derivatives of linear PDEs \cite{LeGuen2020} and a cross-entropy loss, which evaluates the ability of the transformer to predict the next patch encoding.

\begin{equation}\label{phytransformer}
    \mathcal{L}_{Phy-transformer} = \mathcal{L}_{moment}+ \mathcal{L}_{cross-entropy} \\
\end{equation}
with 
\begin{equation}\label{moment}
    \mathcal{L}_{moment} = \sum_{i \leq k}\sum_{j \leq k}  \| M(w^{k}_{p,i,j}) - \Delta^{k}_{i,j}\|_F \\
\end{equation}
\begin{equation}\label{crossentropy}
    \mathcal{L}_{cross-entropy} = \sum_{i=1}^N\sum_{j=1}^C y_i\log{P_{i,j}}
\end{equation}

where $k$ is size of the convolutional filter, $w_p$ is a parameter of the Phycell, $M(w^p_{k,i,j})$ is the moment matrix, $F$ stands for the Frobenius norm, $\Delta^{k}_{i,j}$ is the target moment matrix. For cross-entropy loss, $N$ is the number of samples, and $C$ is the number of class, $y_i$ is the ground truth class label for the token, and $P_{i,j}$ represents the predicted probability of sample $i$ belonging to class $j$.

\paragraph{Training Details}
We used a similar training setup to \citet{Yan2021} for VideoGPT. Our innovation is incorporating PhyCell into the transformer and disentangling latent tokens modeling into two branches, i.e., the modeling of physics dynamics and the modeling of fine-grain details. To deal with the potential information leakage from the PhyCell to the transformer, the current transformer token encodings only get added to the corresponding PhyCell encoding of the previous time-step. Note that otherwise, the transformer would have access to the PhyCell encoding of the image which it is generating. VQ-VAE is first trained on the entire image sequences from time $t-15$ to $t+15$ with a 2-minute interval (i.e., 8 historical frames and 8 future frames). We then use the trained VQ-VAE to encode video data to latent sequences as training data for the Phy-transformer, along with the encodings of the conditional frames obtained by passing through a 3D ResNet for cross-attention during the training. Both components of SkyGPT, i.e., VQ-VAE and Phy-transformer, are implemented using the deep learning framework PyTorch 1.8.1 and trained on a GPU cluster with an Nvidia A100 (40GB memory) card. %More hyper-parameters details can be found in Appendix \ref{appendixB:model_training_details}.

%\paragraph{Evaluation Metrics}

\subsection{PV Output Prediction}
\label{subsubsec:unet_model}
The PV power output prediction task is the second task in our sequential process. It learns a mapping from the sky image to concurrent PV power output. Such a mapping can be trained on historical real-world images and then applied to our generated future sky images. An analogy one can think of is the computer vision task of estimating the age of people based on their facial images \cite{angulu2018age}.

%\begin{equation}
%{f_p}: \mathcal{I}_{t} \mapsto %\hat{\mathcal{{P}}}_{t}
%\end{equation}

\paragraph{Model Architecture} 

\begin{figure}[h!]
    \centering
    \includegraphics[width=.8\textwidth]{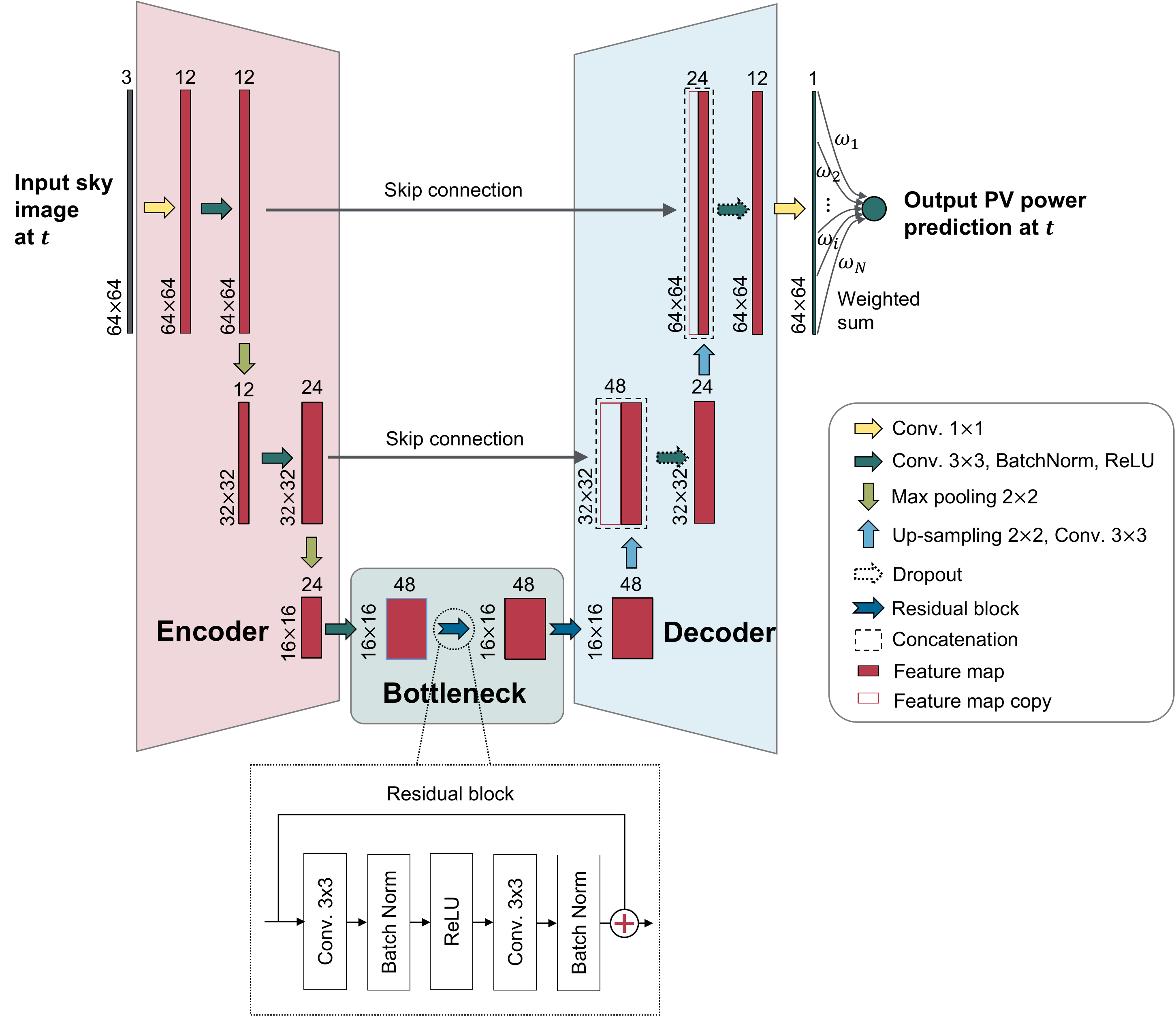}
    \caption{Modified U-Net architecture for PV output prediction. For presentation purposes, we combine the height and width dimensions for the output feature maps, shown as $height \times width$.}
    \label{fig:unet_architecture}
\end{figure}

The PV output predictor is based on U-Net \cite{ronneberger2015u}, which has an encoder-bottleneck-decoder architecture and is commonly used in various image segmentation tasks. For the PV output prediction task, a few modifications were made to the architecture of U-Net, including (1) changing the output of the original U-Net to generate a regression result instead of a segmentation map, (2) using residual block for the bottleneck part instead of the classical Convolution-BatchNorm-ReLU structure to ease the network training, (3) pruning the architecture by reducing the number of convolution layers. The architecture designs of the modified U-Net are shown in Figure \ref{fig:unet_architecture}. The encoder part is composed of a series of 2D convolutions to compress the high-dimension input image data into a low-dimension latent. The latents then pass through the residual blocks and get upsampled and convolved to the same resolution as the input via the decoder, which consists of a series of up-sampling followed by 2D convolutions. However, instead of reconstructing the input, a feature map with the same height and width as the input image but with only one channel is produced and regressed to a single PV output value. This feature map can be viewed as a map of PV output values associated with each pixel and the PV output prediction equals the weighted sum over them. Not only did we find this improved accuracy, but it also makes the model substantially more interpretable. Further, skip connections have been added to pass features from the encoder to the decoder in order to recover spatial information lost during downsampling.

\paragraph{Objective Function}
The modified U-Net is trained deterministically by minimizing mean squared error (MSE) between the ground-truth PV output and the predictions:
\begin{equation}
    \mathrm{MSE} = \frac{1}{N}\sum_{i=1}^N(\hat{\mathcal{P}}_i - \mathcal{P}_i)^2
    \label{eq:MSE}
\end{equation}
where $N$ is the number of samples, $\hat{\mathcal{P}}_i$ is the prediction generated by the model and $\mathcal{P}_i$ is the ground-truth measurement.

\paragraph{Training and Inference} 
The data used for training the modified U-Net are pairs of real sky images and concurrent PV output measurement. Adam \cite{kingma2014adam} is used as the optimizer and a scheduled learning rate decay is applied, which follows the equation below:

\begin{equation}
    lr = lr_0\times \gamma^{\lfloor\frac{epoch}{10}\rfloor}
\end{equation}

Where $lr_0 = 2\times 10^{-4}$ is the initial learning rate to start training, $\gamma=0.5$ is a parameter that controls the rate of decay, $epoch$ stands for the number of training epoch, $\lfloor\ \rfloor$ is the floor function which returns the greatest integer less than or equal to the input argument. We trained the model with 10-fold cross-validation to avoid over-optimistic estimation of the model performance, so essentially 10 sub-models are obtained. The model was coded using the deep learning framework TensorFlow 2.4.1 and trained on a GPU cluster with an Nvidia A100 (40GB memory) card. 
%\paragraph{Inference}

The model is trained deterministically. In order to generate a range prediction of PV output 15 minutes ahead during the inference phase, a collection of possible futures at time $t+15$ generated by the stochastic video prediction model SkyGPT are fed to the modified U-Net model. The number of futures to be generated ($N_f$) is a hyper-parameter, and we conducted experiments to find a balance between model performance and computational cost (see Section \ref{subsec:prob_solar_forecast_results}). Besides, each one of the sub-models from 10-fold cross-validation can generate a PV output prediction; hence, ten predictions can be obtained per each input image. For our proposed probabilistic solar forecasting framework, with the $N_f$ generated realizations of the future sky, a total of $10\times N_f$ predictions can be generated to form the PV output prediction range for a given forecasting time.

\subsection{Baseline Solar Forecasting Framework}
\label{subsec:baseline_solar_forecasting_framework}
\begin{figure}[h!]
    \centering
    \includegraphics[width=.55\textwidth]{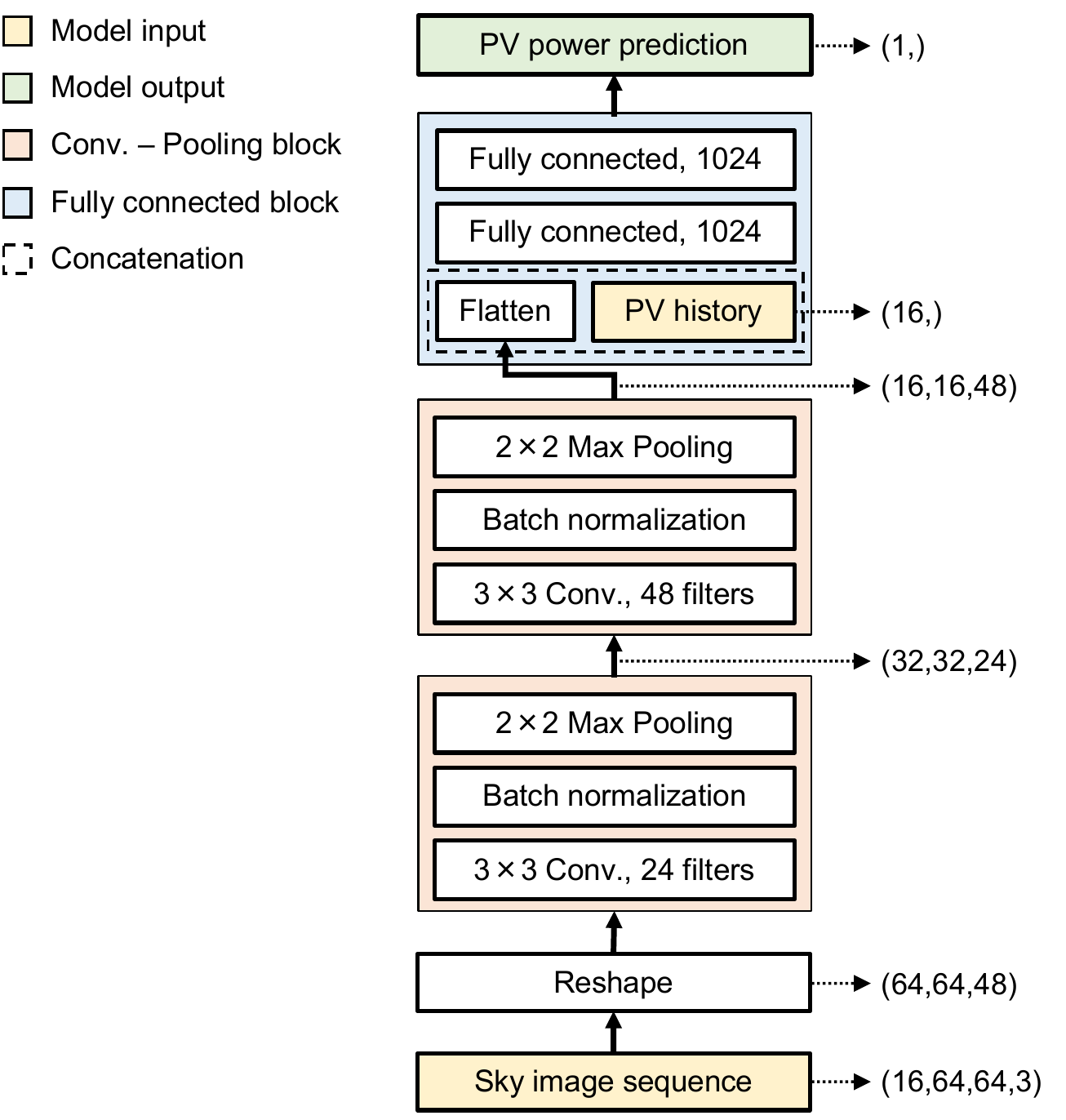}
    \caption{The SUNSET forecast model architecture, with image sequences entering at the bottom, going through layers of operation, and forming an output. 3$\times$3 Conv., 24 filters denotes a convolutional layer with 24 filters, and each filter with a size of 3$\times$3. Fully connected, 1024 denotes a fully connected layer with 1024 neurons.}
    \label{fig:sunset_architecture}
\end{figure}

For comparison with the proposed framework, the baseline forecasting framework considered is the SUNSET model developed by \citet{Sun2019}. The SUNSET model \footnote{SUNSET is open-sourced, and the code base can be accessed \url{https://github.com/YuchiSun/SUNSET} for a TensorFlow 1.X version or \url{https://github.com/yuhao-nie/Stanford-solar-forecasting-dataset} for a TensorFlow 2.x version.} is a CNN-based model that takes in a hybrid input of sky image sequence and concurrent PV output measurement to predict the future PV power generation. The architecture of SUNSET forecast model is shown in Figure \ref{fig:sunset_architecture}. The SUNSET is trained using the same setup described in \citet{Sun2019}. It should be noted that SUNSET is a deterministic prediction model. For consistent comparison with the proposed probabilistic forecasting framework, we use 10-fold cross-validation to train 10 sub-models so that a range prediction can be generated. %The basic structure includes two convolution–pooling (Conv.-Pooling) blocks and one fully connected block. Each convolution–pooling block includes sequentially one convolutional layer, one batch normalization layer, and one max pooling layer. The convolutional layer uses a filter size of 3$\times$3, with a stride of 1 and same-value padding. The activation function used here is a rectified linear unit (ReLU). A 2$\times$2 max pooling with stride of 2 is used in the pooling layer. The first Convolution–Pool block contains 24 filters, while the second contains 48 filters. After the two convolution–pooling blocks, the processed input is flattened, concatenated with PV output history, and passed through two fully connected layers, each containing 1024 neurons. ReLU is used as the activation function and dropout, with a 0.4 dropout rate, is performed to prevent over-fitting. A final regression step with linear weights is used to produce the predicted PV output. 

%\begin{equation}
%f_F:(\mathcal{I},\mathcal{P})_{t-T:\delta:t} \mapsto \mathcal{P}_{t+T}
%\end{equation}

\subsection{Performance Evaluation}
\label{subsec:eval_metrics}
\paragraph{Video Prediction} The predicted image frames are evaluated both qualitatively based on human perception and quantitatively via commonly used evaluation metrics for benchmarking video prediction models, including mean squared error (MSE), mean absolute error (MAE) and VGG cosine similarity (VGG CS) \cite{Lee2018}. For definitions of these metrics, here we define $\mathcal{I} \in \mathbb{R}^{H\times W\times C}$ as the real image, $\hat{\mathcal{I}} \in \mathbb{R}^{H\times W\times C}$ as the predicted image, where $H, W, C$ represents the height, width and number of channels of the images, respectively. 
\begin{equation}
    \mathrm{MSE} = \frac{1}{N}\sum_k^N\sum_i^{H\times W\times C} (\mathcal{I}_i^k-\hat{\mathcal{I}}_i^k)^2
\end{equation}

\begin{equation}
    \mathrm{MAE} = \frac{1}{N}\sum_k^N\sum_i^{H\times W\times C} |\mathcal{I}_i^k-\hat{\mathcal{I}}_i^k|
\end{equation}

\begin{equation}
    \mathrm{VGG\ CS} = \frac{1}{N}\sum_k^N(\frac{\bm{v_k}\cdot \bm{\hat{v}_k}}{||\bm{v_k}||||\bm{\hat{v}_k}||})
    \label{eq:vgg_sim}
\end{equation}
Where $\bm{v_k}=\mathrm{VGG}(\mathcal{I}_k)$, $\bm{v_k}=\mathrm{VGG}(\hat{\mathcal{I}}_k)$, VGG stands for the VGG16 network \cite{simonyan2014very} used to extracted features from the input images. We used the VGG16 model pre-trained on ImageNet from TensorFlow Keras API. The feature vectors used here are drawn from the output of the second to last max pooling layer of the VGG. It should  be noted that VGG16 network requires an input size of $224\times 224$, we therefore resized our input images from $64\times 64$ to $224\times 224$ using Python OpenCV library for evaluating VGG CS.

Although MSE and MAE are widely used metrics for video prediction evaluation, they are not necessarily indicative of prediction quality. For example, models can produce blurry predictions in order to minimize MSE as the loss function. These blurry predictions might not look good to human perception, but they can have good score in terms of MSE or MAE. VGG CS, to this end, is reported to align better with human perception \cite{Lee2018}. We compute and compare these metrics based on samples from the validation set (see Section \ref{sec:dataset} for details of the experimental dataset). These quantitative metrics are used for evaluating deterministic prediction performance. For stochastic video prediction models, We generated 10 possible futures for each video sample and identify the ``best'' sample based on VGG CS to the ground truth video for computing these metrics. 

\paragraph{PV Output Prediction} Reliability and sharpness are two important properties of probabilistic forecasts. Reliability indicates how similar the distribution of the forecast is to that of the observation and sharpness refers to the concentration of the predictive distribution. A good probabilistic prediction thus should have both reliability to cover the observation and sharpness to be informative. In this study, we evaluate the PV output prediction performance based on two probabilistic forecasting metrics --- continuous ranked probability score (CRPS) \cite{gneiting2014probabilistic} and Winkler score (WS) \cite{winkler1972decision}. Both CRPS and WS allows for simultaneous assessment of reliability and sharpness. The CRPS measures the difference between the cumulative distribution function of observation ($F_{target}$) and model prediction ($F_{model}$) as shown in Equation \ref{eq:CRPS}, and the lower the CRPS, the better the prediction is.
%\textcolor{red}{a plot showing it is necessary to use both two metrics. A uniform distribution between 0-30 which results in a WS of 30, but relatively high CRPS. A plot showing same CRPS but different WS value.}

\begin{equation}\label{eq:CRPS}
    \mathrm{CRPS} = \frac{1}{N}\sum_{k=1}^N\int_{-\infty}^{+\infty} \bigr[F_{target}^k(x)-F^k_{model}(x)\bigr]^2dx
\end{equation}

$F_{target}$ is a form of the Heaviside step function, which jumps from 0 to 1 at the value of the measured PV output. $F_{model}$ is the cumulative distribution of the PV output predictions generated by the model. The CRPS is computed as an average over a set of $N$ predictions. An advantage of the CRPS is that it is dimensionally the same as the prediction target
(kW for the PV output in this study) and it reduces to the absolute error if the forecast is deterministic, and therefore allows for comparison between probabilistic and point forecasts \cite{gneiting2014probabilistic}. 

WS is defined in Equation \ref{eq:winkler_score} with a nominal confidence level $(1-\alpha)\%$:

\begin{align}
\label{eq:winkler_score}
    \mathrm{WS}_k & = \begin{cases} 
      \delta & \mathrm{if}\ L_k\leq x_k\leq U_k \\
      \delta+2(L_k-x_k)/\alpha &  \mathrm{if}\ x_k<L_k \\
      \delta+2(x_k-U_k)/\alpha &  \mathrm{if}\ x_k>U_k
   \end{cases}
\end{align}

where $\delta=U_k-L_k$ with $U_k$ and $L_k$ representing the lower and upper bounds of the prediction interval. In this study, we use 95th percentile and 5th percentile of the predictions for $U_k$ and $L_k$, respectively, resulting in a nominal confidence level of 90\% (i.e., $\alpha=0.1$). The WS increases when the observation ($x_k$) lies outside the prediction interval and a wide prediction interval will also be penalized even if it covers the observation; therefore, a lower WS represents a better probabilistic forecast. In order to assess the overall performance, the average WS is calculated over a set of N predictions:

\begin{equation}
    \mathrm{WS} = \frac{1}{N}\sum_k^N \mathrm{WS}_k
\end{equation}

The proposed model is also compared with other benchmark models. A widely established method for assessing the performance of different models is to assess their performance relative to a baseline model. The resulting indicator is called forecast skill (FS). FS essentially quantifies how much better/worse the error of the model is compared to that of a reference model:

\begin{equation}
    \mathrm{FS} = \left(1-\frac{\mathrm{Error}_{forecast}}{\mathrm{Error}_{ref}}\right)\times 100\%
    \label{eq:FS}
\end{equation}

The mostly used reference model in solar forecasting is the smart persistence model (SPM), which assumes the relative output, measured as the ratio of the actual PV output  to the theoretical PV output under clear sky conditions, stays constant from time $t$ to ($t+T$):

\begin{equation}
    k_{clr} = \frac{\mathcal{P}(t+T)}{\mathcal{P}_{clr}(t+T)} = \frac{\mathcal{P}(t)}{\mathcal{P}_{clr}(t)}
    \label{eq:persistence_model}
\end{equation}

where $k_{clr}$ represents the relative output, or formally named as clear sky index, $\mathcal{P}$ is the actual PV output, and $\mathcal{P}_{clr}$ is the theoretical PV output. At any given time stamp, $\mathcal{P}_{clr}$ can be estimated by a clear sky model based on sun angles and PV panel orientations \cite{DaRosa2009}:

\begin{equation}
P_{clr}(t)=P_mA_e\{\cos{\epsilon}\cos{\chi(t)} + \sin{\epsilon}\sin{\chi(t)}\cos[\xi(t)-\zeta]\}
\label{eq:clear_sky_model}
\end{equation}

Where $P_{m}$ is the maximum solar irradiance, 1000 $\mathrm{W/m^2}$; $A_e$ is the effective PV panel area, 24.98 $\mathrm{m^2}$, which is obtained from a least square fit with the real panel output of 12 clear sky days (details can be found in study by \citet{Sun2019}); $\epsilon$ and $\zeta$ are elevation and azimuth angles of the solar PV arrays, which are $22.5^{\circ}$ and $195^{\circ}$, respectively; $\chi(t)$ and $\xi(t)$ are the zenith and azimuth angle of the sun, which can be estimated for any minute of the year from the empirical functions provided in the textbook by \citet{DaRosa2009}.

Based on Equation \ref{eq:clear_sky_model}, $T$-minute-ahead PV output can be estimated by SPM:

\begin{equation}
    \hat{\mathcal{P}}(t+T)=\frac{\mathcal{P}(t)}{\mathcal{P}_{clr}(t)}\times \mathcal{P}_{clr}(t+T)
    \label{eq:persistence_pred}
\end{equation}

Here, the error metric we used for calculating FS is CRPS. For SPM, as it can only generate 1 prediction each time, the CRPS is reduced to mean absolute error (MAE) \cite{gneiting2014probabilistic}:

\begin{equation}
    \mathrm{MAE} = \frac{1}{N}\sum_{i=1}^N|\hat{\mathcal{P}}_i - \mathcal{P}_i|
    \label{eq:MAE}
\end{equation}

\section{Dataset}
\label{sec:dataset}
\paragraph{Overview} We leverage an in-house dataset\footnote{This study was conducted before the official release of our curated dataset SKIPP'D \cite{nie2022skipp}, which is more organized and has a number of updates from the dataset we used here. We encourage the readers to examine the SKIPP'D dataset \url{https://github.com/yuhao-nie/Stanford-solar-forecasting-dataset}.} ($\mathscr{D}$) with 334,038 aligned pairs of sky images ($\mathcal{I}$) and PV power generation ($\mathcal{P}$) records, $\mathscr{D} = \{(\mathcal{I}_i, \mathcal{P}_i) \mid i\in \mathbb{Z}: 1\leq i\leq 334\mathrm{,}038\}$, for the experiments in this study. The sky image frames are extracted with a 1-minute frequency from video footage recorded by a ground-based fish-eye camera (Hikvision DS-2CD6362F-IV) installed on the roof of the Green Earth Sciences Building at Stanford University. The images are down-scaled from 2048$\times$2048 to 64$\times$64 pixels to save model training time. The PV power generation data are collected from a 30-kW rooftop PV system $\sim$125 meters away from the camera, with an elevation angle of 22.5\degree and an azimuth angle of 195\degree \footnote{The azimuth angle is measured clockwise between the North and the PV panel orientation.}. PV data are minutely averaged and paired with the image data according to the time stamps. The collection period of the dataset is from March 2017 to December 2019, with some disruptions because of the water intrusion, wiring and/or electrical failure of the camera as well as a daylight-saving adjustment.

\paragraph{Data Processing} As this study aims to address the challenges of solar forecasting on cloudy conditions, we use the cloudy samples from $\mathscr{D}$ for model development and test. We focus on cloudy conditions because they correspond to the times when PV prediction is nontrivial -- predicting clear sky PV output is a well-understood problem \cite{peratikou2022estimating}. To filter out the clear sky samples, we follow the algorithm developed by \citet{Nie2020}, which detects cloud pixels in sky images based on a modified normalized red blue ratio method. This screening step results in a cloudy sample subset ($\mathscr{D}_{\mathrm{cloudy}}$) consisting of 132,305 samples from $\mathscr{D}$. 

For different forecasting tasks described in Section \ref{sec:methodology}, i.e., cloud motion prediction and PV output prediction, different model input and output configurations are used, thus requiring different organization of samples. A common processing step is carried out to form an interim dataset ($\mathscr{D}_{\mathrm{interim}}$), which is then sampled to get the dataset for the specific forecasting task. To obtain $\mathscr{D}_{\mathrm{interim}}$, we loop through the time stamps in $\mathscr{D}_{\mathrm{cloudy}}$ with a step size of 2 minutes (a so-called 2-minute sampling frequency by \citet{Sun2019}) to check if the future (and historical) PV output and sky image records from next (last) minute to next (past) 15 minutes are available with 1-minute resolution. The sampling frequency is chosen to be 2 minutes because a higher frequency can lead to longer model training time with limited improvement on the model accuracy \cite{Sun2019}. Any sample that does not satisfy the above conditions is filtered out. After this processing step, 60,385 valid samples are obtained, $\mathscr{D}_{\mathrm{interim}}=\{(\mathcal{I}_{t-15:1:t+15}^i, \mathcal{P}_{t-15:1:t+15}^i) \mid i\in \mathbb{Z}: 1\leq i\leq \mathrm{60,385}\} \}$, with each sample containing a sequence of 31 sky images ($\{\mathcal{I}_{t-15},\mathcal{I}_{t-14},...,\mathcal{I}_{t}, \mathcal{I}_{t+1}, ..., \mathcal{I}_{t+15}\}$) and PV output measurements ($\{\mathcal{P}_{t-15},\mathcal{P}_{t-14},...,\mathcal{P}_{t}, \mathcal{P}_{t+1}, ..., \mathcal{P}_{t+15}\}$). 

To form the dataset for the video prediction task ($\mathscr{D}_{\mathrm{vp}}$), only the image data from $\mathscr{D}_{\mathrm{interim}}$ is used. For each image sample, we apply a 2-minute interval to save the model training time as the video prediction task is computationally expensive. This results in $\mathscr{D}_{\mathrm{vp}}=\{(\mathcal{I}_{t-15:2:t-1}^i, \mathcal{I}_{t+1:2:t+15}^i) \mid i\in \mathbb{Z}: 1\leq i\leq \mathrm{60,385}\} \}$, where the first 8 images (from $t-15$ to $t-1$) can be used as model input while the remaining 8 images (from $t+1$ to $t+15$) can serve as prediction target. For the PV output prediction task, we have two different models, our proposed model modified U-Net and baseline model SUNSET. For the U-Net model, we take both image and PV output data at time $t$ for each sample of $\mathscr{D}_{\mathrm{interim}}$, forming the U-Net dataset $\mathscr{D}_{\mathrm{unet}}=\{(\mathcal{I}_{t}^i, \mathcal{P}_{t}^i) \mid i\in \mathbb{Z}: 1\leq i\leq \mathrm{60,385}\} \}$ where image $\mathcal{I}_{t}^i$ is served as model input and PV output $\mathcal{P}_{t}^i$ is served as model output. While for the SUNSET model, we take images and PV output records from time $t-15$ to time $t$ with 1-minute resolution, and PV output at time $t+15$ for each sample of $\mathscr{D}_{\mathrm{interim}}$, forming the SUNSET dataset $\mathscr{D}_{\mathrm{sunset}}=\{[(\mathcal{I}_{t-15:1:t}^i, \mathcal{P}_{t-15:1:t}^i), \mathcal{P}_{t+15}^i] \mid i\in \mathbb{Z}: 1\leq i\leq \mathrm{60,385}\} \}$. The image and PV output sequence $(\mathcal{I}_{t-15:1:t}^i, \mathcal{P}_{t-15:1:t}^i)$ is served as the model input and PV output $\mathcal{P}_{t+15}^i$ is served as model output. %It should be noted that all three datasets contain the same amount of samples as $\mathscr{D}_{\mathrm{interim}}$.

%After screening, we obtain 53,336 (88\%), 4,467(7\%) and 2,582 (4\%) valid samples, respectively, for the forecast problems. $57\mathrm{,}803$ aligned sample pairs $\{c_{t-T:\delta:t}^i, f_{t+1:\delta:t+T}^i\}$, where $i\in \mathbb{Z}: 1\leq i\leq 57\mathrm{,}803$, $T=15$ and $\delta=1$. We separate 10 cloudy days with totally $4\mathrm{,}467$ samples from the dataset for model testing, and the remaining samples are used for model development. For the sake of saving computations in training, we down-sampled the image frames to a resolution of $64\times64$ and take 2-minute time interval between image frames, i.e., $\delta=2$, so each $c_{t-15:2:t-1}^i$ and $f_{t+1:2:t+15}^i$ sample is a $8\times64\times64\times3$ tensor. 

The sky images used in this study can be segmented into two parts, namely, the area within the circle showing the view of the camera (referred to as the foreground), and the black area outside the circle (referred to as the background). Although the background looks black, the pixel values are not necessarily all 0s, but actually could be some small values. Our initial experiments have found that these pixels contain some useful information for PV output forecasting task, and we guess that they might be caused by some sort of backscattering within the camera sensor, although it needs to be further validated in the future. However, for this study, as we want our models to focus on the foreground to make predictions, we mask out the background by applying a binary mask with all 0s for pixels of the background and all 1s for pixels of the foreground to the original images pixel-wise. In the following two cases, the binary mask is applied: (1) For evaluating the performance of video prediction models, the images generated by the video prediction models are masked. Note that the mask is not applied to the images used as input of the video prediction model for training. (2) For PV output prediction model training, validation and testing, either real images or images generated by the video prediction models are masked.

\paragraph{Data Partitioning} The processed samples are then split into training, %($\mathscr{D}_{\mathrm{train}}$), 
validation %($\mathscr{D}_{\mathrm{val}}$) 
and testing set. %($\mathscr{D}_{\mathrm{test}}$) 
We manually select 10 cloudy days from 2017 March to 2019 October based on their PV output profile for the validation set, taking into account the seasonal and annual variations. The rest of the data from the same period (i.e., 2017 March to 2019 October) goes to the training set. To test the model generalizability, we select 5 cloudy days from 2019 November to December to form the testing set, which is outside the time window of training and validation data. Samples from the validation and testing sets are not included in the model training process to prevent over-optimistic model estimation. The above split results in 53,336 (88\%), 4,467(7\%) and 2,582 (4\%) samples, for the training, validation and testing set, respectively, for the forecast tasks. The video prediction model is only trained and validated without evaluating on the test set. The PV output prediction model uses all of the three datasets. 

\section{Results}
\label{sec:experiments}

\subsection{Sky Video Prediction}
%We first experiment with different architecture designs and hyper-parameters to identify the optimal framework for sky image prediction (see supplementary materials for details). The identified best performing model uses a transformer combined with a PhyCell. Figure \ref{fig:generated_img_demo} shows examples of sky image prediction based on two different context frames via our best-performing model. As a comparison, we also show the prediction by benchmark models for spatiotemporal sequence forecasting, including ConvLSTM, PredRNN, PredRNNv2, PhyDNet, VideoGPT. It shows that compared with these benchmark models, our model can generate more realistic future cloud images with less blurriness and capture the cloud dynamics in the context frames better. Then we focus on quantitative analysis of the fidelity and diversity of the predicted images from our proposed model, and we further apply the generated images for a downstream solar forecasting task and assess the performance by RMSE of the PV output prediction.

Although video prediction is not the ultimate goal of this study, understanding how well the predicted images align with the ground truth helps facilitate the analysis of the errors of the solar forecasting system. As a comparison, a few open-source benchmark video prediction models were trained based on our dataset, including ConvLSTM\cite{Shi2015}, PhyDNet \cite{LeGuen2020} and VideoGPT \cite{Yan2021}. ConvLSTM and PhyDNet are deterministic prediction models, while VideoGPT is stochastic. GANs are a commonly used architecture for various generation tasks and are especially known for their power in generating very high-fidelity images. In this study, we also implemented adversarial training based on the PhyDNet architecture (referred to as PhyDNet+GAN), which is also deterministic. 

The performance of the video prediction models is evaluated using the samples from the validation set, both qualitatively based on human perceptual judgments and quantitatively based on the performance metrics described in Section \ref{subsec:eval_metrics}. Two aspects are considered for assessing the generated images, namely, realism and diversity. On the one hand, we want the predicted sky images to be as realistic as possible and close to the real sky images. On the other hand, given the uncertainty in cloud motion, we want the generated images to be reasonably diverse to cover different possible scenarios of the future sky. The generated samples by deterministic models are only evaluated for realism, while those by the stochastic models are evaluated for both realism and diversity.

%We first show qualitative results of our proposed SkyGPT model on the validation set by comparing with the benchmark models in Figure \ref{fig:qual_eval_benchmark_comp}. 

%\begin{figure*}[h]
%\vskip 0.2in
%\begin{center}
%\centerline{\includegraphics[width=\textwidth]{Figures/qual_eval_benchmark_comp_1.pdf}}
%\caption{Examples of qualitative evaluation of the predicted frames by SkyGPT compared to benchmark models. The left part shows an example with increasing amount of cloud and the right part shows an example of decreasing amount of cloud. The first row represents the context frames from $(t-15)$ to $(t-1)$ as input to the model. The second row shows the ground truth future images from $(t+1)$ to $(t+15)$. The rest of rows demonstrate the predicted future frames from our Proposed model SkyGPT and benchmark models.}
%\label{fig:qual_eval_benchmark_comp}
%\end{center}
%\vskip -0.2in
%\end{figure*}

\begin{figure}[h!]
     \centering
     \begin{subfigure}[b]{1.0\textwidth}
         \centering
         \includegraphics[width=\textwidth]{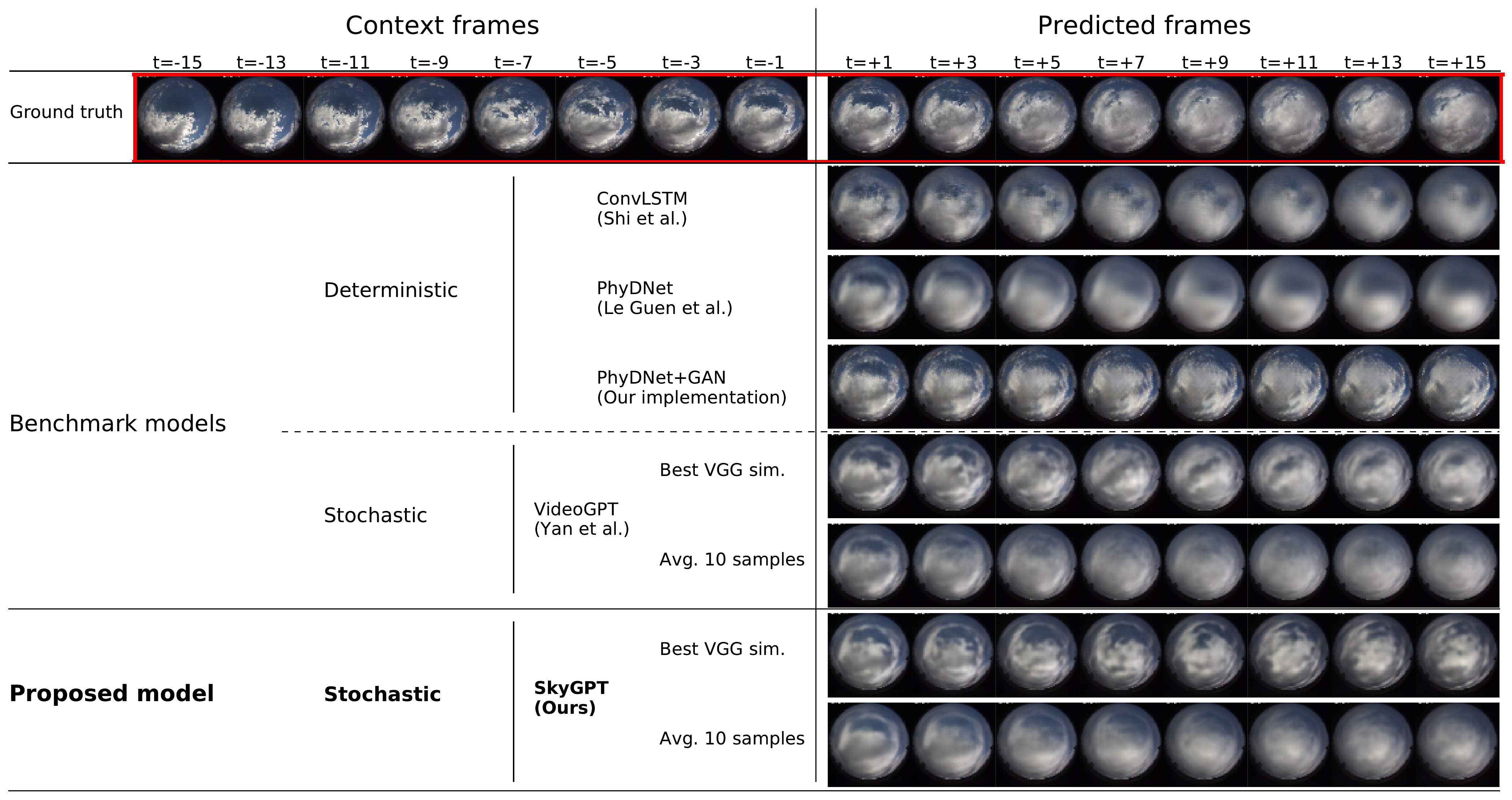}
         \caption{An example of the sky changing from partly cloudy to overcast condition}
         \label{fig:demo_partly_cloudy_to_overcast}
     \end{subfigure}
     \hfill
     \begin{subfigure}[b]{1.0\textwidth}
         \centering
         \includegraphics[width=\textwidth]{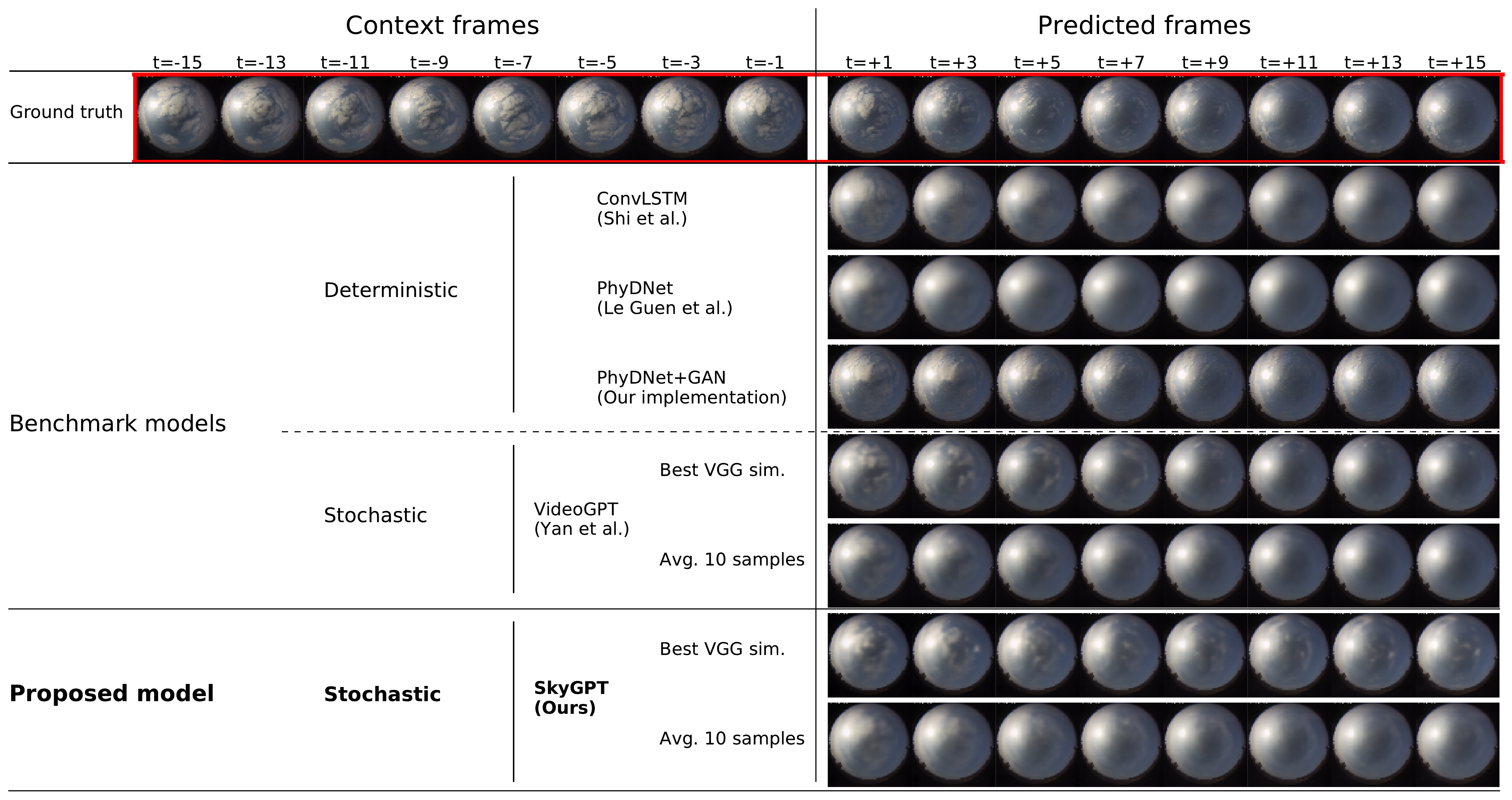}
         \caption{An example of the sky changing from partly cloudy to clear sky condition}
         \label{fig:demo_partly_cloudy_to_clear_sky}
     \end{subfigure}
        \caption{Examples of the predicted frames by SkyGPT compared to benchmark models. The left part of the figure represents the historical (context) frames as input to the model. The right part shows the future frames, with the first row representing the real sky (frames highlighted in red) and the rest representing the predicted sky by different models.}
        \label{fig:demo_qual_results}
\end{figure}

\paragraph{Realism of the Prediction}
\label{qual_eval}
%Figure \ref{fig:demo_qual_results} shows two samples generated by our proposed SkyGPT model on the validation set as well as the benchmark models. It can be observed that our proposed model tend to generate much clear results compared with the deterministic benchmark models, which looks blurry.
Figure \ref{fig:demo_qual_results} shows predictions from the proposed SkyGPT model as well as benchmark video prediction models based on two sets of historical inputs. These two examples illustrate two different cloud dynamics: (a) the sky changing from partly cloudy to overcast condition and (b) the sky changing from partly cloudy to clear sky condition. For the stochastic prediction models, i.e., VideoGPT and SkyGPT, ten future samplings were generated and two cases are shown here, the most similar generation among the ten generations to the real future images as measured by VGG CS (referred to as Best VGG sim.), and the pixel average of the ten generations (refer to as Avg. 10 samples).  

For both examples, most video prediction models could capture the general trend of cloud motion correctly, except that PhyDNet fails to capture the correct dynamics for example (a). Although all model predictions show different extents of flaws, VideoGPT and our proposed model SkyGPT seem to capture the cloud dynamics better than other benchmark models. Note that the bright spot in example (a) is moving from the right bottom corner to the middle and only VideoGPT and SkyGPT captures it correctly. SkyGPT is a bit better than VideoGPT in terms of overall light and shading. For the general appearance, deterministic models ConvLSTM and PhyDNet could generate clear images for the immediate future but beyond a few frames they start to produce blurry predictions. By training the same PhyDNet architecture with a GAN framework (PhyDNet+GAN), it is able to generate the clearest images among all models even when moving into the far future. However, the texture, light, and shading of the generations is not better than VideoGPT and SkyGPT. In comparison, the predicted frames from these two stochastic models look much more clear than ConvLSTM and PhyDNet, and are competitive with the prediction generated by PhyDNet+GAN. The images look less blurry for the far future as both models use a transformer for prediction, which is proven to work well for long-term sequence modeling. 

\begin{figure}[h!]
    \centering
    \includegraphics[width=1.0\textwidth]{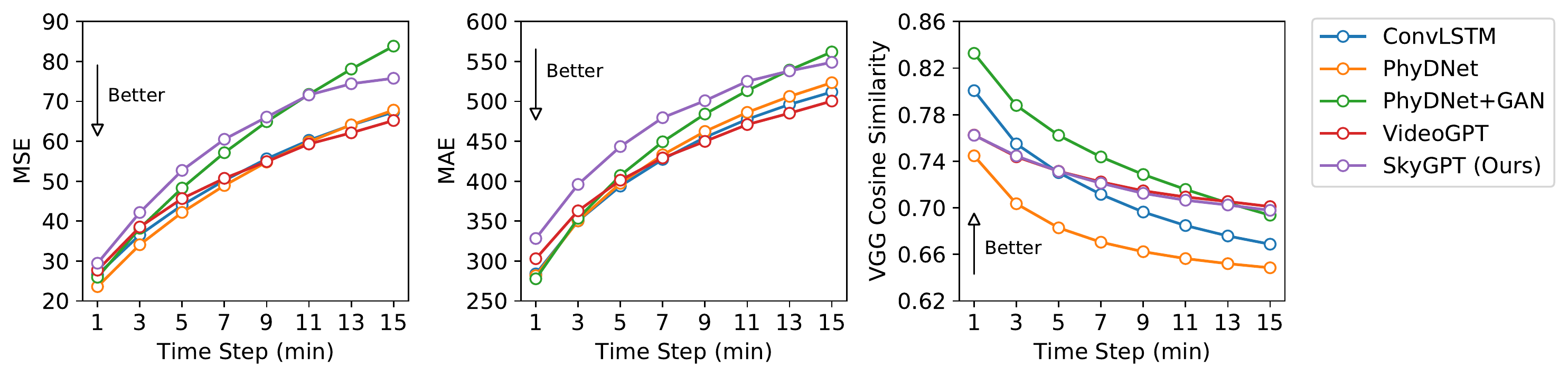}
    \caption{Quantitative evaluation of SkyGPT compared to benchmark video prediction models with different time steps}
    \label{fig:quan_eval_results}
\end{figure}

Figure \ref{fig:quan_eval_results} shows the quantitative evaluation results of the predicted images of all video prediction models based on three metrics, i.e., MSE, MAE and VGG CS. It should be noted that for the stochastic models VideoGPT and SkyGPT, the results shown here reflect the average of the best performance over the 10 samplings (i.e., the minimum MSE and MAE, and the maximum VGG CS). Generally, we observe a trend of performance degradation of the predicted images over time, regardless of the metrics. Although the benchmark models ConVLSTM and PhyDNet generate images with more blurriness, they tend to have better performance in MSE and MAE relative to PhyDNet+GAN and SkyGPT, which generate images with higher fidelity. This is expected as ConvLSTM and PhyDNet try to minimize the MSE/MAE of the images during the training process, and produce blurry predictions as a way of accommodating the uncertainty in cloud motion. Similar results are observed by \citet{Lee2018}. In terms of VGG CS, which is more aligned with human perception, images with more realism generated by models PhyDNet+GAN, VideoGPT, and SkyGPT have higher scores especially when it evolves into the far future. Note that when the timestep is greater than 5 minutes, these models can outperform ConvLSTM and PhyDNet. 

%\paragraph{Fidelity} 

\paragraph{Diversity of the Prediction}
\label{quan_eval}

\begin{figure*}[htbp!]
\vskip 0.2in
\begin{center}
\centerline{\includegraphics[width=1.0\textwidth]{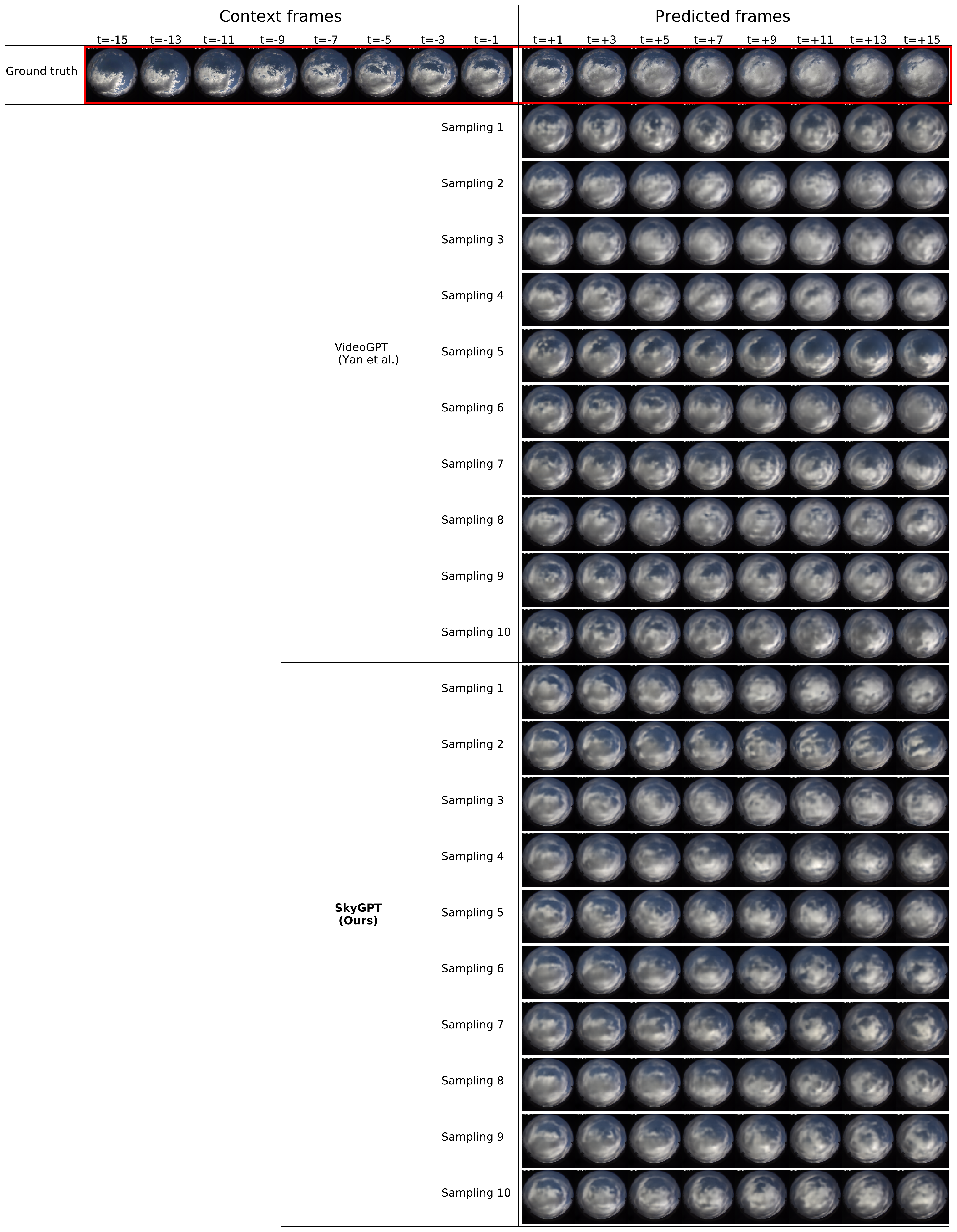}}
\caption{Demonstration of the diversified predictions from the VideoGPT and our proposed SkyGPT model. Here, we show 10 different possible future predictions generated by each model based on the same context frames. The first row represents the real sky (frames highlighted in red) and the rest of rows represent the predicted sky.}
\label{fig:model_diversity_eval}
\end{center}
\vskip -0.2in
\end{figure*}

We first assess the diversity of the samples generated by the two stochastic video prediction models based on visual inspection. Specifically, we used the same historical frames as model input and generate ten different futures and check how different they are. Figure \ref{fig:model_diversity_eval} shows the ten samples generated by VideoGPT and the proposed model SkyGPT, respectively. It can be observed that frames generated for the immediate future could be similar, but the generations start to diverge from each other beyond a few frames. Note the similarity between different samplings for the first two future frames at $t=+1$ and $t=+3$ for SkyGPT and the diversity for the rest of the future frames. Similar findings can be observed for Sampling 5-10 generated by VideoGPT. Another point to be made here is that some cloud motion patterns might be generated more frequently than others by the models. According to the ground truth sample, the clouds likely move from the bottom left to the top right. Most of the generations from SkyGPT and VideoGPT successfully capture this dynamic although the details of the generations could be varied, e.g., the cloud coverage, the light and shading and texture of clouds. An exception is found for Sampling 5 generated by VideoGPT, which clearly has different dynamics, i.e., the clouds change moving direction after a few time steps. This is not impossible, as the wind direction could change, but this might be less likely based on the dynamics of the historical frames. In comparison, SkyGPT seems to consistently capture similar motion dynamics as the ground truth sample, probably due to the addition of the physics-constrained module PhyCell in its architecture.

\begin{figure}[htpb!]
    \centering
    \includegraphics[width=.8\textwidth]{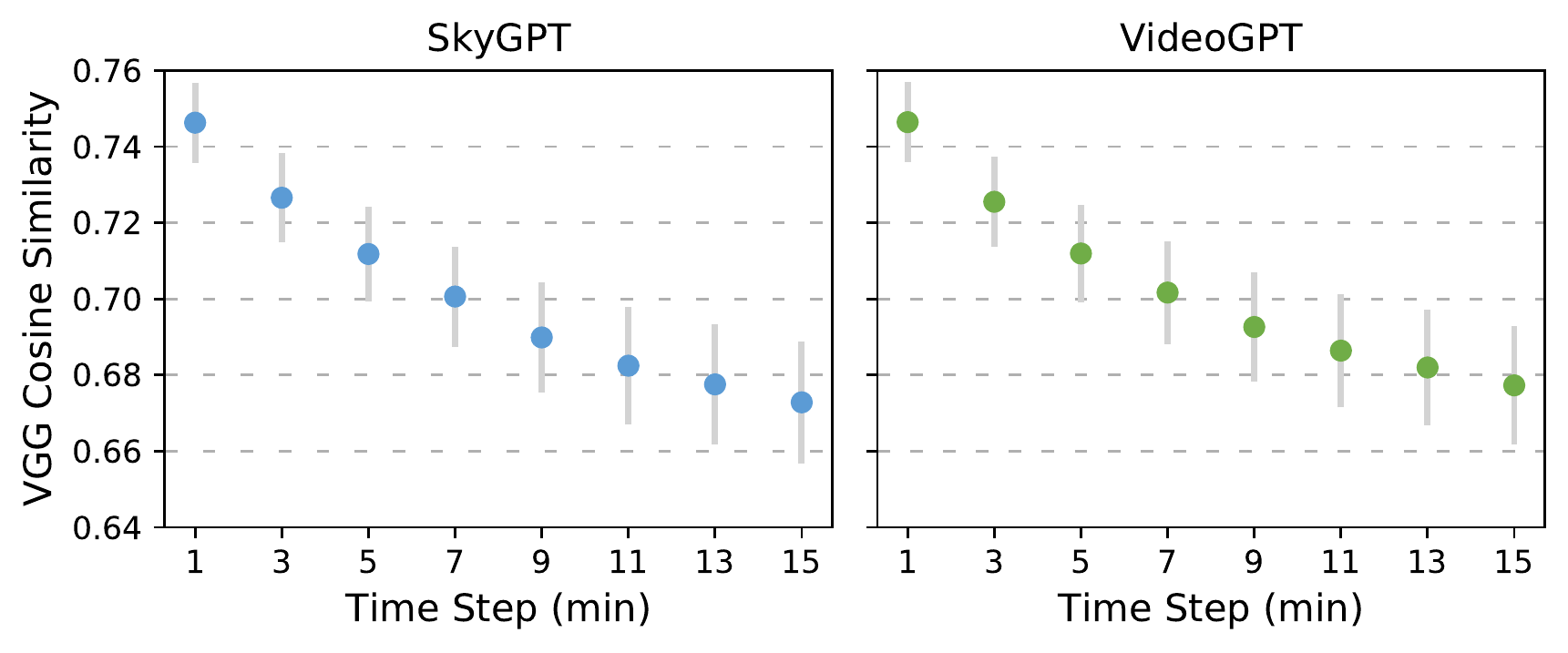}
    \caption{Diversity of the generated samples from the stochastic video prediction models SkyGPT and VideoGPT. The dots represent the mean VGG CS of 10 samples and the error bar shows one standard deviation. Both mean and standard deviation are averaged over all samples in the validation set. Note the increase of the error bar length from the immediate future to the far future.}
    \label{fig:quan_diversity}
\end{figure}

The diversity of the generated samples are also evaluated quantitatively. We generated ten different futures for every sample in the validation set, computed the mean and standard deviation of the VGG CS of these ten future samplings relative to the ground truth, and averaged them over all samples in the validation set for each time step. The results are shown in Figure \ref{fig:quan_diversity}, where the dots stand for the mean VGG CS and the error bar represents one standard deviation. It should be highlighted here that the standard deviation can reflect how these ten future samplings differ from each other, i.e., a larger value means the ten generated futures are more diverse from each other. Both SkyGPT and VideoGPT show an increase of the standard deviation over time, indicating the future frames gradually diverge from the immediate future to the far future. This finding is consistent with the results shown in Figure \ref{fig:model_diversity_eval}. Degradation of mean VGG CS over time is also observed, which is similar to the trend observed in Figure \ref{fig:quan_eval_results} although the method of calculating the VGG CS is slightly different. For VGG CS results shown in Figure \ref{fig:quan_eval_results}, it is calculated based on the maximum VGG CS of ten different future generations (i.e., find the one future that is the closest to the ground truth in terms of the VGG CS) for each sample and averaged over the whole validation set. 

\subsection{Probabilistic Solar Forecasting}
\label{subsec:prob_solar_forecast_results}
As a second step, we apply the predicted sky images from the video prediction models for a probabilistic short-term solar forecasting task which aims at forecasting the 15-min-ahead power output of a 30 kW PV system (descriptions of the experiment data can be found in Section \ref{sec:dataset}). The range of PV output prediction at time $t+15$ is derived from multiple point predictions that are generated by feeding each of the predicted images at time $t+15$ from the video prediction models to the modified U-Net model, which is trained on the real sky images and PV output pairs (see Section \ref{subsubsec:unet_model}). For the deterministic video prediction models, i.e., ConvLSTM, PhyDNet and PhyDNet+GAN, although only one future is generated for each input historical sky image sequence, 10 PV output predictions are generated based on the ten U-Net sub-models from 10-fold cross-validation conducted during training. For the stochastic video prediction models, i.e., VideoGPT and SkyGPT, 10 possible futures are generated for each historical sky image sequence and fed to 10 U-Net sub-models, thus, 100 predictions in total are generated. %It should be noted for the base case, the generated images are directly injected into the U-Net model for PV output inference. Experiments on fine-tuning the pre-trained U-Net model with the generated images are presented later in this section. Unless otherwise stated, the results shown are associated with U-Net model without fine-tuning.

\paragraph{Comparison of Different Forecasting Methods} 
For comparing with the performance of using the generated images from these video prediction models for PV output prediction, the following three baseline solar forecasting methods are considered: (1) the SUNSET forecast model adapted from \citet{Sun2019} as described in Section \ref{subsec:baseline_solar_forecasting_framework}, which takes in a hybrid input of sky images and PV output history from the past 15 minutes; (2) the smart persistence model (see Equations \ref{eq:persistence_model} to \ref{eq:persistence_pred} in Section \ref{subsec:eval_metrics}), a commonly used reference model in the solar forecasting community, which assumes the relative power output (the ratio between real power output and power output under clear sky condition) is preserved for the 15 min forecasting horizon; and (3) predictions generated by feeding the real future sky images at $t+15$ into the same U-Net model, which is a hypothetical case that demonstrates the forecasting system's performance if the video prediction component were 100\% accurate. This value represents an upper bound on performance. It should be noted that the smart persistence model can only generate one point prediction at a given forecasting time stamp, while the SUNSET model and U-Net model fed with real future sky image can generate 10 predictions due to 10-fold cross-validation conducted during training. 

The models are evaluated on the same validation and test set (see Section \ref{sec:dataset}) and the probabilistic forecasting performance measured by CRPS, FS, and WS for all methods are presented in Table \ref{tab:solar_forecasting_performance}. The baseline SUNSET shows 13\% and 10\% FS relative to the smart persistence reference for validation and testing, respectively. Feeding predicted frames from video prediction models to the PV output prediction model U-Net, we indeed see benefits in terms of both CRPS and WS. Compared with smart persistence model, 6\% to 18\% and 18\% to 23\% FS can be achieved for the validation and the testing set, respectively. Especially for testing test, respectively, the studied two-stage forecasting methods show better generalization compared with SUNSET. Among the two-stage frameworks, the performance varies with specific video prediction models utilized for generating future images. Methods using stochastic video prediction models generally outperform those with deterministic video prediction models in terms of both CRPS and WS. It is expected as stochastic video prediction models can generate multiple possible futures, which have more chance to cover the real future sky conditions, as indicated by the significantly better WS. Our proposed method SkyGPT$\rightarrow$U-Net consistently performs well for both validation and testing set. However, we still have a gap compared to the case where the U-Net is applied to true future images, which suggests a need for further improvement of the video prediction model. 

\begin{table}[htpb!]
\caption{Solar forecasting methods performance on the validation and test set measured by CRPS, FS and WS. Smart persistence, SUNSET, and real future sky image (SI)$\rightarrow$U-Net are baselines for comparison, and the rest are the studied two-stage forecasting methods. Experimented methods with the best performance are highlighted in bold font.} %ConvLSTM, PhyDNet, PhyDNet+GAN are deterministic video prediction models and VideoGPT and SkyGPT are stochastic video prediction models. For each forecasting time stamp, smart persistence model generates only 1 PV output prediction, SUNSET, real future SI$\rightarrow$U-Net, and deterministic video prediction model$\rightarrow$U-Net generate 10 PV output predictions, and stochastic video prediction model$\rightarrow$U-Net generate 100 PV output predictions.}
\label{tab:solar_forecasting_performance}
\begin{center}
\begin{tabular}{lcccc}
\toprule
\multirow{2}{*}{Solar forecasting method}  & \multicolumn{2}{c}{Validation} & \multicolumn{2}{c}{Test} \\ \cmidrule(lr){2-3} \cmidrule(lr){4-5}
   & CRPS [kW] $\downarrow$  (FS [\%] $\uparrow$) & WS $\downarrow$ & CRPS [kW] $\downarrow$ (FS [\%] $\uparrow$) & WS $\downarrow$\\ \midrule
 Smart persistence & 2.89 (0\%) & - & 3.67 (0\%)& -\\ 
SUNSET & 2.51 (13\%) & 42.73 & 3.31 (10\%) & 56.95 \\
Real future SI$\rightarrow$U-Net & 1.71 (41\%) & 20.91 & 2.34 (36\%) & 27.13 \\
\midrule
ConvLSTM$\rightarrow$U-Net & 2.70 (7\%) & 41.12 & 3.09 (16\%) & 43.82\\
PhyDNet$\rightarrow$U-Net & 2.72 (6\%) & 41.28 & 3.12(15\%) & 45.20 \\
PhyDNet+GAN$\rightarrow$U-Net & 2.50 (14\%) & 36.23 & 3.02 (18\%) & 39.54 \\
VideoGPT$\rightarrow$U-Net & \textbf{2.37 (18\%)} & \textbf{25.20} &  3.34 (9\%) & 36.22 \\
\textbf{SkyGPT$\rightarrow$U-Net (Proposed)} & 2.49 (14\%) & 26.72 & \textbf{2.81 (23\%)} & \textbf{26.70} \\
\bottomrule

\end{tabular}
\end{center}
\end{table}

\begin{figure}[h!]
    \centering
    \includegraphics[width=1.\textwidth]{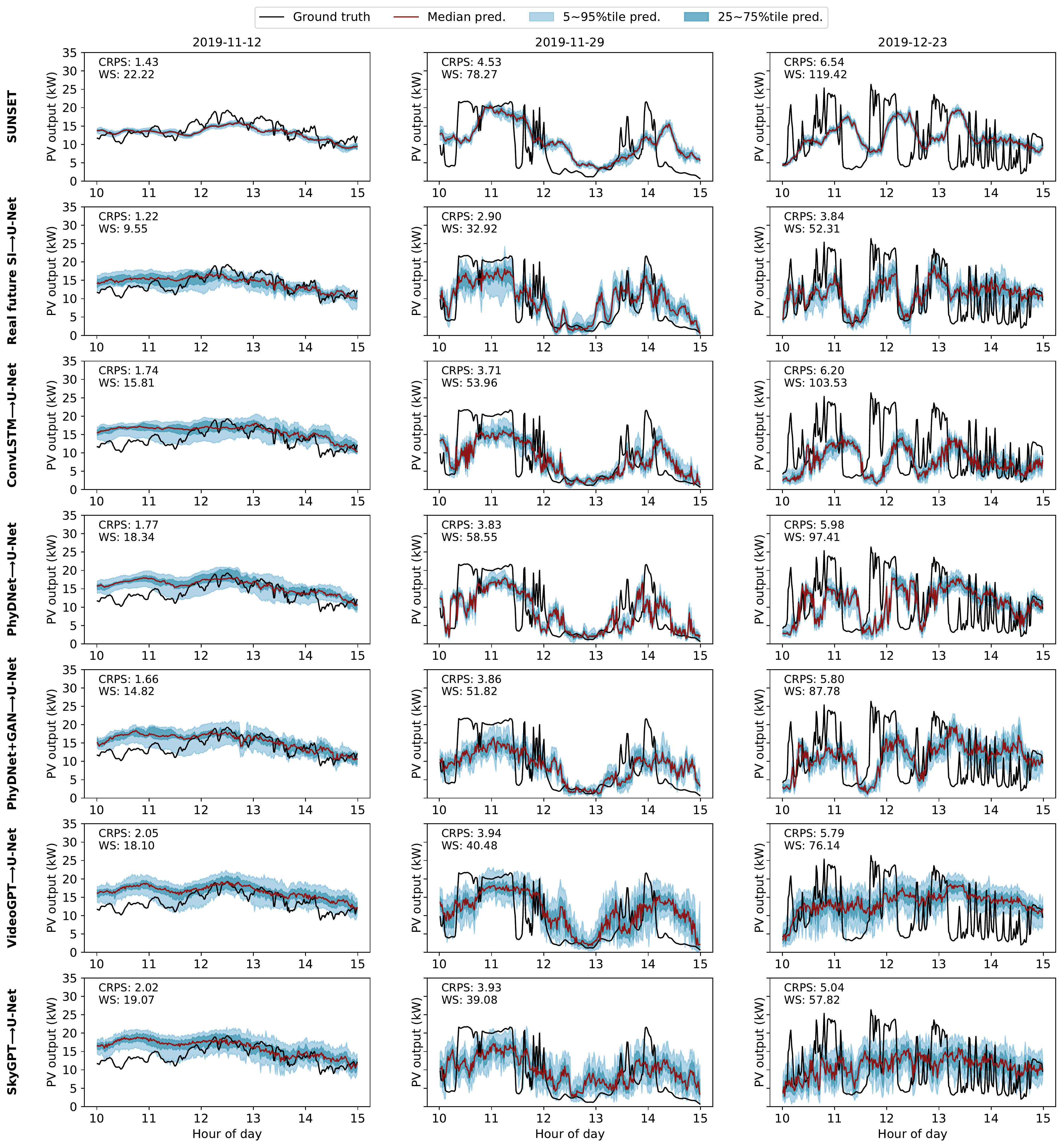}
    \caption{Prediction curves for three cloudy days for different forecasting methods.}
    \label{fig:prediction_curves_3_cloudy_days}
\end{figure}

Figure \ref{fig:prediction_curves_3_cloudy_days} shows the prediction curves for different forecasting methods for 3 cloudy days in the test set for the time period 10:00 to 15:00. The 3 days show increasing variation of PV output as indicated by the spikes and drops of the curves. In the figure, we show two intervals of the predictions, one is 5 to 95 percentile in light blue shade (indicating we are 90\% confident that the measurement will fall in this range), and another is 25 to 75 percentile in the dark blue shade (indicating we are 50\% confident that the measurements will fall in this range). The baseline SUNSET model has a hard time capturing the ramp events and shows lags in predicting the spikes and drops, especially on days with highly varied PV output (see 2019-12-23). Besides, the prediction intervals of the SUNSET model tend to be consistently narrow, which is probably due to the fact that the model relies heavily on the 16 historical images and PV output records to make the predictions, and the variations between the 10 sub-models are limited. In comparison, the U-Net frameworks that directly correlate feature images with PV output generally show larger variations regardless of the video prediction models utilized, indicating that the 10 sub-models are more diverse compared with SUNSET. Deterministic video prediction coupled with the U-Net framework shows less satisfactory performance for days with high variations in PV output. The ramp events are not successfully captured by the median prediction nor covered by the prediction intervals. For the stochastic video prediction model coupled with U-Net frameworks, better coverage of the ramp events is observed. The trend of the median predictions tends to fall in the middle. Another observation is that when the PV output is less varied, the prediction intervals are narrow, while when there are lots of variations in PV output, the prediction intervals tend to be wider. This is as to be expected.

\begin{figure}[htpb!]
    \centering
    \makebox[\textwidth][c]{\includegraphics[width=.9\textwidth]{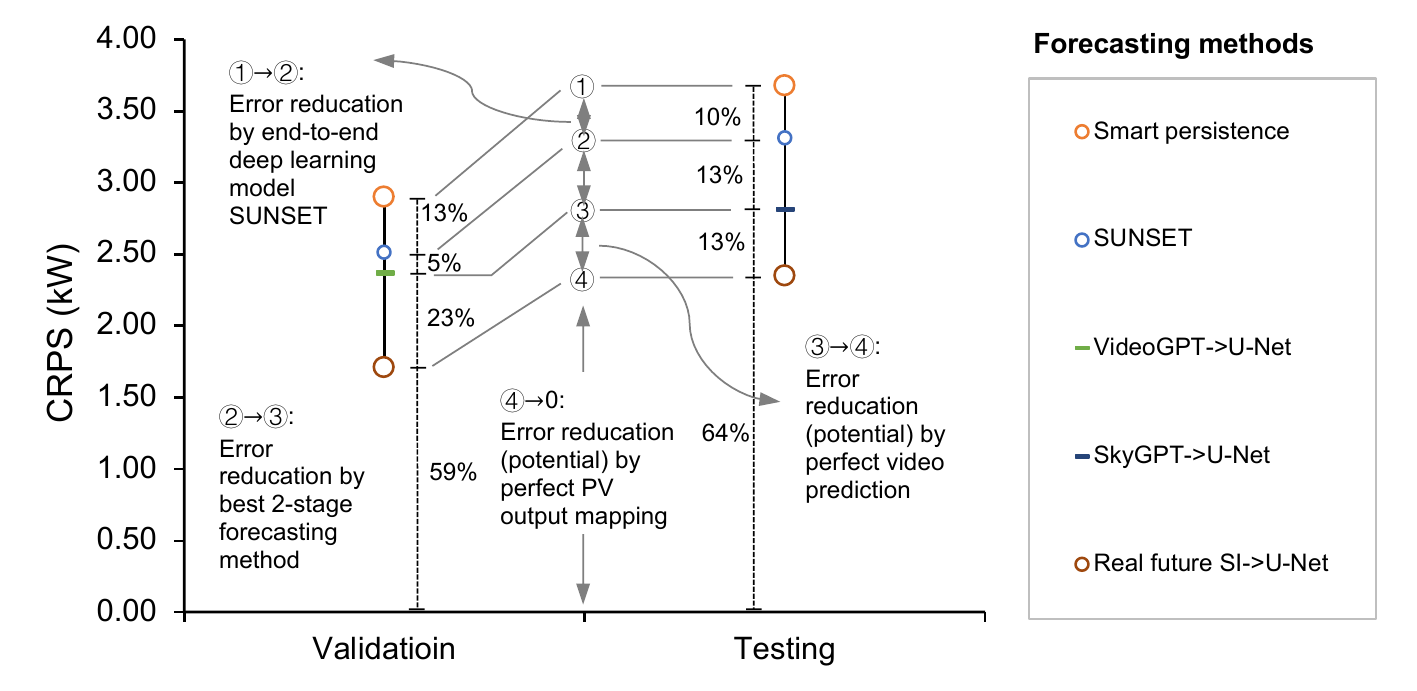}}
    \caption{Comparison of various solar forecasting methods in terms of CRPS for validation and testing sets. The percentages in the figure show the difference in normalized CRPS between different forecasting methods. The normalized CRPS for each forecasting method is calculated based on the CRPS of the smart persistence model (i.e., $\mathrm{CRPS/CRPS_{SPM}}$). For the two-stage forecasting framework, only the best-performing method is shown here, namely, VideoGPT for the validation set and SkyGPT for the test set.}
\label{fig:error_decomposition}
\end{figure}

\paragraph{Error Analysis of the Proposed Forecasting System}
The error of the proposed 2-stage forecasting framework can largely be attributed to two parts, i.e., the prediction of future sky images and the mapping from sky images to contemporaneous PV output. Figure \ref{fig:error_decomposition} compares the performance of different solar forecasting methods on a high level based on the CRPS of the validation and testing sets shown in Table \ref{tab:solar_forecasting_performance}. The percentages in the figure show the difference in normalized CRPS between different forecasting methods, indicating the improvement one can obtain going from one method to another. The normalized CRPS for each forecasting method is calculated based on the CRPS of the smart persistence model (i.e., $\mathrm{CRPS/CRPS_{sp}}$). Generally, going from naive baseline smart persistence model to deep-learning baseline SUNSET, the error can be reduced by 10\%$\sim$13\%, and an additional 5\%$\sim$13\% improvement can be obtained by using the best performing 2-stage forecasting framework proposed in this study, i.e., VideoGPT$\rightarrow$U-Net for the validation set and SkyGPT$\rightarrow$U-Net for the test set. If the real future sky images at time $t+15$ are fed to the U-Net model, the performance can be further boosted by 13\% to 23\%, which shows the potential of improving the video prediction component of the 2-stage forecasting system. However, to further reduce the error of the forecasting system, it relies on the improvement of the PV output mapping model which causes the major error even under the assumption of a perfect video prediction. The residual error at \circled{4}, $\sim$2.3 kW CRPS, represents the error remaining when going from a \emph{true} future sky image to PV output, and represents the error associated with taking an image at time $t$ and producing the associated PV output at that time (the so-called ``now-cast'' problem).

%\paragraph{Effects of Fine-tuning on PV Output Prediction}
%Figure \ref{fig:finetuning_effects} shows the effects of fine-tuning the U-Net model pre-trained with the real image and PV output data. The results show that the fine-tuning has little effect on the probabilistic forecasting metrics CRPS and WS. The models without fine-tuning sometimes even perform slightly better than the ones with fine-tuning. It is probably because the U-Net is trained based on minimizing the RMSE, not CRPS or WS, as evidenced by a slightly improvement on the root mean squared error (RMSE) for the fine-tuning models.

%\begin{figure}[h!]
    %\centering
    %\makebox[\textwidth][c]{\includegraphics[width=1.0\textwidth]{Figures/finetuning_effects.pdf}}
    %\caption{Comparison of PV output prediction performance with fine-tuning versus without fine-tuning.}
%\label{fig:finetuning_effects}
%\end{figure}

\paragraph{Number of Possible Futures to be Generated}

\begin{figure}[htpb!]
    \centering
    \includegraphics[width=.8\textwidth]{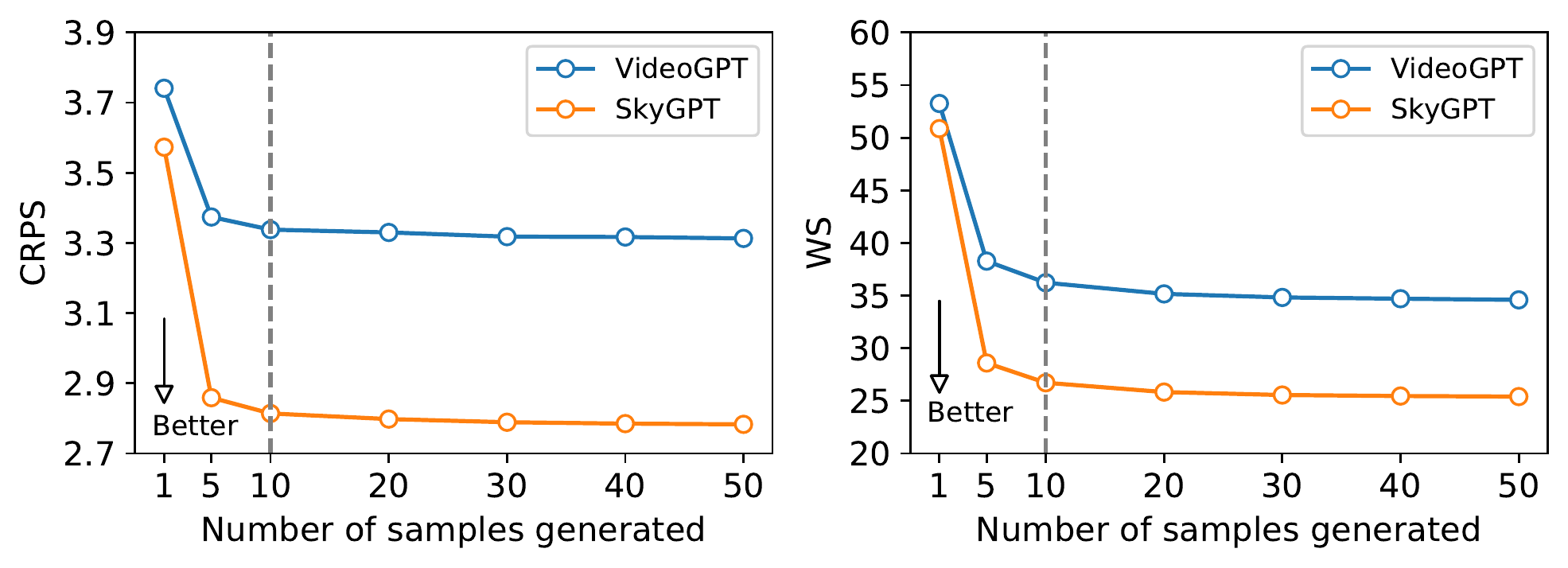}
    \caption{CRPS and WS as a function of number of samples generated by stochastic prediction models (Note number of samples generated equal 1 is equivalent to deterministic video prediction).}
    \label{fig:effect_of_num_of_samps_generated}
\end{figure}

Increasing number of future sky image samples generated by the stochastic model could potentially increase the chance of the real future sky condition being covered. However, it costs more computational effort to generate more future sky images. As a reference, on Stanford's high-performance computing cluster with access to a single Nvidia A100 GPU, it takes roughly 23 hours to generate 50 different future samplings for each of the 2,582 samples in the testing set using the trained SkyGPT (i.e., on average, it takes ~32 seconds to generate 50 future samplings per sample). Therefore, a balance between the performance boost and computational cost must be found. Figure \ref{fig:effect_of_num_of_samps_generated} shows CRPS and WS as a function of number of samples generated by stochastic video prediction models. It shows that the benefits of generating more future samples almost plateau at around 10 samples. Increasing from 1 sample to 10 samples generated, significant reductions in both CRPS and WS are found due to increased coverage of the possible futures by the video prediction models. After 10 samples were generated, the improvement is less significant as there might be more overlap between the newly generated images with the old ones. Therefore, generating 10 different future samplings from the stochastic video prediction models for downstream PV output prediction would be a good option in terms of both performance and computation.

\begin{figure}[htpb!]
    \centering
    \includegraphics[width=.8\textwidth]{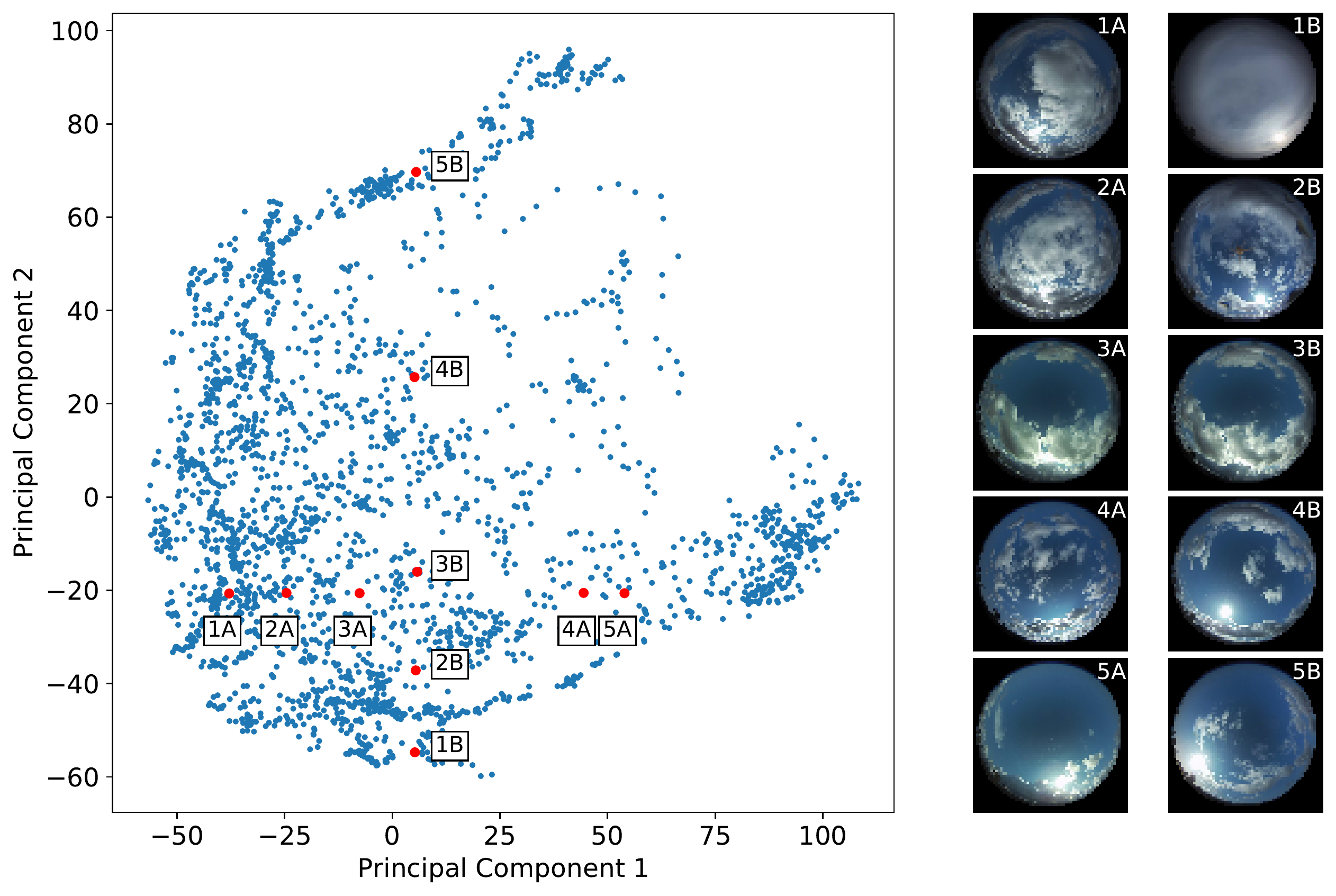}
    \caption{The distribution of the first two principal components of the feature vectors of all 2,582 sky image samples at time $t+15$ in the test set based on PCA. The images on the left-hand side with annotated labels are drawn from the distribution. The left column presents labeled images 1A to 5A (corresponding to the first principal component), showing variability in cloud coverage. The right column presents labeled images 1B to 5B (corresponding to the second principal component), showing differences in horizontal sun position.}
\label{fig:pca_demo}
\end{figure}

\begin{figure}[htpb!]
     \centering
     \begin{subfigure}[b]{1.0\textwidth}
         \centering
         \includegraphics[width=\textwidth]{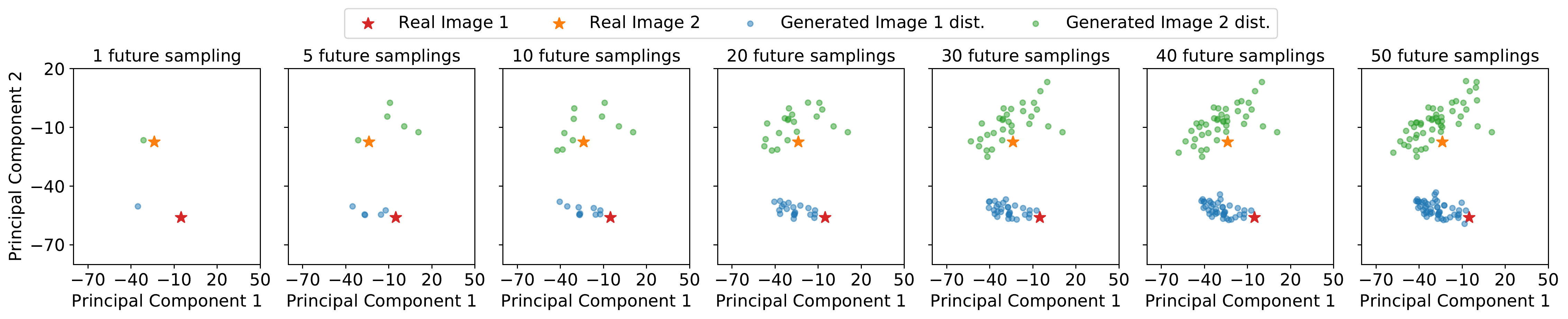}
         \caption{Image distributions for different number of future samplings}
         \label{fig:image_dist}
     \end{subfigure}
     \hfill
     \begin{subfigure}[b]{1.0\textwidth}
         \centering
         \makebox[\textwidth][c]{\includegraphics[width=1.\textwidth]{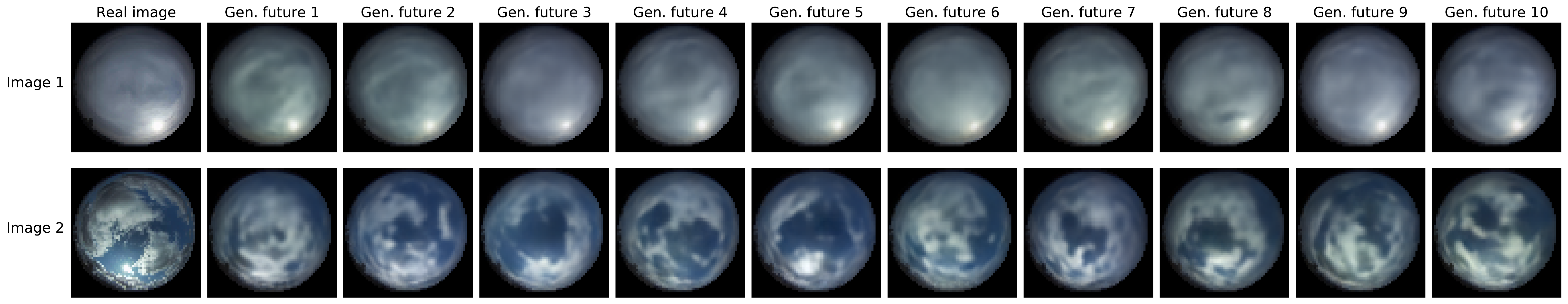}}
         \caption{Illustrative image samples}
         \label{fig:demo_partly_cloudy_to_clear_sky}
     \end{subfigure}
        \caption{The distributions two image samples in the test set and corresponding generated images with different number of future samplings based on the first two PCA components.} \label{fig:image_dist_based_on_PCA}
\end{figure}

We further validated the above findings by visualizing the distribution of the generated images at time $t+15$ by SkyGPT for different numbers of future samplings. The distributions of high-dimensional images can be visualized in a 2-D space by using the first two principal components of the image feature vectors. During the training of U-Net, it learns to extract features from sky images for PV output predictions; hence we can use the trained U-Net model as a feature extractor for sky images. Specifically, we took the output of the bottleneck part of the trained U-Net model and flattened it to get the feature vectors of the sky image. The sky image feature vectors are standardized to zero mean and one standard deviation and the first two principal components are obtained by principal component analysis (PCA). Figure \ref{fig:pca_demo} shows the distribution of the first two principal components of the feature vectors of all 2,582 sky image samples at time $t+15$ in the test set based on PCA. These two components seem to correspond to the variability of cloud coverage and to the horizontal position of the sun in the sky as similarly observed in a previous study \cite{Paletta2021eclipse}. To illustrate this point, Figure \ref{fig:pca_demo} also shows a few labeled image samples drawn from the distribution. The left column shows labeled images 1A to 5A, and the right column shows labeled images 1B to 5B. It can be observed that images 1A to 5A show variability in cloud coverage, while images 1B to 5B show differences in horizontal sun position. With that, we then drew two different image samples in the test set and visualize the distributions of corresponding generated images for different numbers of future samplings based on the first two PCA components in Figure \ref{fig:image_dist_based_on_PCA}. It shows that the distribution of the 10 future samplings could cover different possibilities fairly well, which reinforces our hypothesis of 10 as the optimal number of future samplings above. It can also be observed that the distribution of Image 1 samplings is relatively concentrated, while the distribution of Image 2 samplings is dispersed. That is mainly due to the different levels of uncertainty in cloud motion, and for the cases with high uncertainty, the SkyGPT model tends to generate more diverse futures.

%\textcolor{red}{Image distribution? based on the feature vectors returned by VGG? PCA?}

\section{Limitations and Future Work}
\label{sec:limitations}

While our proposed probabilistic solar forecasting system shows promising performance for 15-minute-ahead PV output prediction, there are limitations that need to be addressed in future work to further improve the prediction reliability and sharpness. Although both video prediction model and PV output prediction model need improvement, according to the error analysis (see Figure \ref{fig:error_decomposition}), we suggest that the main efforts should be focused on improving the PV output prediction model, i.e., mapping sky images to contemporaneous PV outputs, or the ``nowcast'' problem. 

For prediction reliability, it can be noticed that even we feed the real future sky images to the PV output prediction model, the prediction intervals could sometimes fail to cover the real PV output, especially in peak and valley regions of the PV output curve (see the prediction curve by "Real future SI$\rightarrow$U-Net" on day 2019-12-23 in Figure \ref{fig:prediction_curves_3_cloudy_days}). Here, we provide two potential research directions to ameliorate this:

\begin{enumerate}
    \item Increasing the size and diversity of the training data. In this study, we used only 53,336 samples for training the SkyGPT model for video prediction and the modified U-Net model for PV output prediction, which could be limited. Especially due to SkyGPT's transformer component, it often requires a massive amount of data for training to achieve good performance. Besides, the dataset was solely collected from Stanford campus in California, where cloudy days are normally a lot less than sunny days over the course of a year. Therefore, there are limitations in the amount as well as the diversity of cloudy samples. \citet{nie2022open} recently conducted a comprehensive survey and identified 72 existing open-source sky image datasets collected globally for short-term solar forecasting and related research. It can potentially provide useful information for compiling a large-scale dataset for training the proposed probabilistic solar forecasting system.
    \item Examining emerging architecture for PV output prediction and tuning the hyper-parameters of the video prediction model SkyGPT. We only examined U-Net as the backbone of PV output prediction model in this study, which is essentially a CNN. With the recent advancement in deep learning, transformers has been successfully applied in various computer vision tasks and show competitive  performance as previously dominant CNNs \cite{pinto2022impartial,islam2022recent,zhao2021battle}. Vision transformer (ViT) \cite{dosovitskiy2020image} has great potential to be adapted to the PV output prediction task studied here.  While SkyGPT is a novel combination of VideoGPT and PhyDNet for video prediction, we mostly used the default settings of these two networks for training SkyGPT. Architecture and hyper-parameter tuning of SkyGPT could be helpful in further improving its performance.
    
\end{enumerate}

For prediction sharpness, our proposed system could sometimes generate undesired wide prediction intervals even when there is not much variance in PV output (see time periods 12:30-13:00 and 14:30-15:00 of the prediction curve by "SkyGPT$\rightarrow$U-Net" on day 2019-11-29 in Figure \ref{fig:prediction_curves_3_cloudy_days}). This behavior mainly results from the diverse futures generated by SkyGPT. The possible futures are sampled from the learned distribution that is an approximate to that of the training video dataset without taking into account the corresponding PV output, as the two models are trained separately. A potential way to improve the prediction sharpness is to use the signal from the  PV output prediction model training to guide the stochastic image prediction, specifically, a new loss function can be designed to reward the future image generations that lead to PV output predictions within a certain range around the real PV output while penalizing the image generations that lead to PV output outside this range. In other words, the video prediction model can learn a distribution with controlled variance based on the feedback from the PV output prediction model training.

Another issue that needs attention is the gap between the generated images and the real images. As the PV output prediction model U-Net was trained based on the pairs of real sky image and PV output but never seen generated images during the training process, feeding the generated images to the PV output prediction model could cause errors during the inference phase. We touched base on this issue by applying a strategy called fine-tuning, same as that used by \citet{nie2022sky} for transferring the knowledge from a pre-trained solar forecasting model to a new model. Specifically, we kept the same U-Net architecture for PV output prediction, and initialized the model with the weights from the U-Net model pre-trained based on the real data and further trained it by feeding the generated images with a lower learning rate, thus adjusted new weights could be learned that incorporate information about generated images. We then applied these fine-tuned model to the test set for PV output prediction. However, the results show that the fine-tuning has little effect on the probabilistic forecasting metrics CRPS and WS, and the models without fine-tuning even perform slightly better than the ones with fine-tuning for most of the time. Methods should be developed in the future to deal with the discrepancy in input images for training and inference. %At a given time stamp, multiple possible futures could be generated by the stochastic video prediction model. However, at the same time stamp, there is only one ground-truth PV output label. To handle this issue, we fed all predicted sky images at that time to the pre-trained U-Net model and generate a prediction for each of these images. Then a mean PV output prediction is computed based on these individual predictions and used to minimize the training loss MSE. After the modified U-Net is trained on the real data, we feed it with the generated images (together with the concurrent real PV output measurement) from the validation dataset, which has not been seen during training, to let it adapt to the generated image space.

\section{Conclusion}
\label{sec:conclusion}

This work explores 15-minute-ahead probabilistic PV output prediction for a 30-kW rooftop PV system using synthetic future sky images from deep generative models. We introduced SkyGPT, a stochastic physics-informed video prediction model that is capable of generating multiple possible futures of the sky from historical sky image sequences. Extensive experiments were performed to compare SkyGPT with other benchmark video prediction models for future sky image prediction as well as using the generated images for probabilistic solar forecasting. For video prediction, the qualitative and quantitative results show that the proposed SkyGPT model can effectively capture the cloud dynamics and generate realistic yet remarkably diverse future sky images. It excels in long-term frame prediction and outperforms most of the benchmark models in terms of VGG cosine similarity beyond 5 minutes. For PV output prediction, the collection of generated future images from various video prediction models is fed to a specialized U-Net model and compared to an end-to-end deep learning baseline SUNSET as well as the smart persistence model. Coupling SkyGPT with U-Net shows better prediction reliability and sharpness for the test set than all other solar forecasting methods, achieving a continuous ranked probability score of 2.81 (13\% better than SUNSET and 23\% better than smart persistence) and a Winkler score of 26.70. Error analysis indicates that although video prediction plays an important role in determining the probabilistic forecasting performance, error caused by PV output inference still dominates, further suggesting that more efforts should be put on improving the PV output prediction in future work.

%Some possible directions for the future work are listed here: 
%(1) expand the training dataset to cover different possible cloud patterns. 
% (3) examine forecasting future image in an iterative fashion, e.g., forecasting sky image frames iteratively from $t_0+1$ to $t_0+15$ instead of directly forecasting sky image frame at $t_0+15$.

%\appendix
%\section{Model Architecture Details}
%\label{appendixA:model_arch_details}

%\section{Model Training Details}
%\label{appendixB:model_training_details}

\bibliographystyle{unsrtnat}
\bibliography{references}  

\end{document}